\documentclass[accepted]{uai2026}

\usepackage[american]{babel}
\usepackage{natbib}
\usepackage{mathtools}
\usepackage{amssymb} % for \mathbb and additional symbols
\usepackage{booktabs}
\usepackage{array}
\usepackage{tikz}
\usepackage{float}
\usetikzlibrary{positioning,arrows.meta,fit}

\usepackage{graphicx}
\usepackage{xurl} % allow line breaks in long URLs
\usepackage{wrapfig}
\usepackage{grffile} % allow spaces in file names (e.g., Obsidian paste names)

\title{Gaussian Process Latent Factor Regression for Low-Data Problems with High-Dimensional Outputs}

\author[1]{\href{mailto:es833@cam.ac.uk}{Edward~T.~Stevenson}}
\author[2]{Eric~T.~Wolf}
\author[3]{Mei~Ting~Mak}
\author[4]{N.~J.~Mayne}
\author[1]{Miles~Cranmer}
\affil[1]{University of Cambridge}
\affil[2]{University of Colorado Boulder}
\affil[3]{University of Oxford}
\affil[4]{University of Exeter}

\makeatletter
\renewcommand{\maketitlehooka}{%
  \vbox to 2.875in\bgroup
}
\makeatother

\begin{document}
% --- PREPRINT ONLY (remove \pagestyle{numbered} and \thispagestyle{numbered}
% before camera-ready PDF). The class uses \pagestyle{empty} for [accepted]
% until startpage is given; we override so arXiv shows page numbers without a
% fake proceedings footer.
\pagestyle{numbered}
\maketitle
\thispagestyle{numbered} % undo \maketitle's plain page (empty footer in accepted mode)

\begin{abstract}
  In the sciences, regression tasks often require predicting high-dimensional outputs from few training examples.
  Multi-output Gaussian processes excel in low-data regimes but typically struggle with high-dimensional outputs.
  Compress-then-predict pipelines such as PCA-GP (principal component analysis plus Gaussian process regression) handle high dimensionality, but rely on bases optimized for reconstruction rather than prediction.
  To address this gap, we propose a model that represents each output as a linear-Gaussian decoding of a low-dimensional latent state drawn from a Gaussian process prior.
  By analytically marginalizing the decoder weights, we couple compression and prediction in a single objective that scales to high-dimensional outputs.
  We refer to this model as Gaussian process latent factor regression (GPLFR).
  We demonstrate GPLFR by building the first spatially resolved emulator of global climate models for rocky exoplanets.
\end{abstract}

\section{Introduction}\label{sec:intro}

Multi-output regression can be framed around two broad modeling questions: how do the entries of output $\mathbf{y}$ relate to one another, and how does input-space location $\mathbf{x}$ dictate similarity between examples?
With limited data, these two questions compete for modeling capacity.
When outputs are heterogeneously or sparsely observed, sharing information across outputs is essential, and modeling capacity should flow toward output covariance.
When outputs are high-dimensional and structured, a rich output covariance becomes poorly identified, whereas there is more information to constrain a compressed latent representation: each training example provides many output dimensions as repeated ``views'' of the same underlying latent state; here capacity should flow toward learning the similarity structure over $\mathbf{x}$.

Most multi-output GPs (MOGPs; ``co-kriging'' in geostatistics) prioritize sharing information across outputs by learning an explicit coupling structure, while using a shared input kernel to control generalization across $\mathbf{x}$.
For example, the popular \emph{linear model of coregionalization} (LMC) style MOGPs encode output covariance through linear mixing of shared latent GPs (see \citealp{alvarezKernelsVectorValued2012} for a review).
Latent-variable MOGPs (LV-MOGPs) extend this idea by defining similarity between \emph{outputs} via a kernel on per-output latent embeddings rather than directly parameterizing a coregionalization matrix \citep{daiEfficientModeling2017}.
Other approaches in this lineage include higher-order Gaussian process regression \citep{zheScalableHighOrder2019}, which models output correlations through learned latent coordinate features in a tensorized output space with Kronecker inference, and Gaussian process regression networks \citep{wilsonGaussianProcess2011,liScalableGaussian2020}, which generalize LMC by making the decoder weights input-dependent GP functions. 
Latent-factor GP models in this family also appear in unsupervised settings, (e.g., \citealp{liModelingDynamic2019}).
Exact inference in these models scales poorly in both the number of training examples $N$ and the output dimensionality $D_y$ (as $\mathcal{O}((ND_y)^3)$ for an unrestricted LMC), so scalable variants either restrict the model (e.g., orthogonal mixing; \citealp{bruinsmaScalableExact2020}) or approximate inference, most commonly with sparse variational methods \citep{nguyenCollaborativeMultioutput2014,jiangScalableMultiOutput2025}.
However, even with these extensions, they remain ill-suited to problems where the bottleneck is learning input-space structure rather than enriching output coupling.

The standard alternative approach is to decouple the problem into two stages: first learn a basis from outputs alone (usually via PCA), then regress the resulting coefficients against $\mathbf{x}$ \citep{hutchingsFastEmulation2025, holdenEmulationInterpretation2015, higdonComputerModel2008, rougierEfficientEmulators2008}.
While this compress-then-predict approach handles high $D_y$, its basis is optimized for output reconstruction rather than predictability from the inputs.
In this work, we avoid this mismatch by learning compression and regression jointly in a way that scales to high $D_y$.\footnote{We discuss adjacent task-aware representation-learning methods in Appendix~\ref{app:adjacent-task-aware}.}
Concretely, our proposed model -- Gaussian process latent factor regression (GPLFR) -- infers latents that are simultaneously constrained by a GP prior over $\mathbf{x}$ and reconstruction of $\mathbf{y}$ through a linear decoder.

\begin{figure*}[t]
  \centering
  \resizebox{0.98\textwidth}{!}{\begingroup\def\GPLFRPGMINMAIN{}% GPLFR plate diagram.
%
% This file is intentionally dual-purpose:
% - When compiled directly, it produces a standalone PDF.
% - When included from MAIN.tex, it emits only the TikZ picture.

\providecommand{\GPLFRPGM}{%
\begin{tikzpicture}[
    >=stealth,
    node distance=1.0cm,
    obs/.style={circle,draw,fill=gray!20,minimum size=26pt, inner sep=1pt},
    latent/.style={circle,draw,minimum size=26pt, inner sep=1pt},
    plate/.style={draw, rounded corners=5pt, draw=black!80},
    every edge/.style={draw, ->, shorten >=1pt}
]

% --- Nodes ---

% Per-latent GP hyperparameters (positioned higher to make q-plate taller)
\node[latent] (ell) at (-0.7, 2.5) {$\boldsymbol{\ell}_q$};
\node[latent] (eta) at (0.7, 2.5) {$\eta_q$};

\node[latent] (z) at (0, 0.0) {$z_i^{(q)}$};

% Observed input/output
\node[obs] (xi) at (-2.5, 0.0) {$\mathbf{x}_i$};
\node[obs] (yi) at (2.5, 0.0) {$\mathbf{y}_i$};

% Decoder globals (balanced around y, and far enough right to avoid plate overlap)
\node[latent] (B) at (4.7, 2.5) {$\mathbf{B}$};
\node[latent] (W) at (4.7, 0.9) {$\mathbf{W}$};
\node[latent] (sigma) at (4.7, -0.9) {$\sigma$};

% --- Plates ---

% Inner Plate (q=1...Dz) - intersects i-plate at z
\node[plate, fit=(ell) (eta) (z), inner xsep=0.30cm, inner ysep=0.30cm,
      label={[inner sep=3pt]above:{\small $q=1,\dots, D_z$}}] (plateQ) {};

% Outer Plate (i=1...N)
\node[plate, fit=(xi) (yi) (z), inner xsep=0.6cm, inner ysep=0.6cm,
      label={[inner sep=3pt]below left:{\small $i=1,\dots,N$}}] (plateN) {};

% --- Edges ---

% Into z
\draw (ell) edge (z);
\draw (eta) edge (z);
\draw (xi) edge (z);

% Into y
\draw (z) edge (yi);
\draw (B) edge (W);
\draw (W) edge (yi);
\draw (sigma) edge (yi);

% --- Info Box ---

\node[draw, rectangle, text width=7.1cm, align=left, right=1.2cm of W, yshift=0.3cm] (infobox) {
  \textbf{Variables:}\\[6pt]
  \hspace{0.8em}$\mathbf{x}_i \in \mathbb{R}^{D_x}$: input\\[4pt]
  \hspace{0.8em}$\mathbf{y}_i \in \mathbb{R}^{D_y}$: output\\[4pt]
  \hspace{0.8em}$\boldsymbol{\ell}_q \in \mathbb{R}_+^{D_x}$: kernel lengthscales\\[4pt]
  \hspace{0.8em}$\eta_q \in \mathbb{R}_+$: kernel amplitude\\[4pt]
  \hspace{0.8em}$z_{i}^{(q)} \in \mathbb{R}$: latent variable\\[4pt]
  \hspace{0.8em}$\mathbf{W} \in \mathbb{R}^{D_y \times D_z}$: decoder weights\\[4pt]
  \hspace{0.8em}$\sigma \in \mathbb{R}_+$: observation noise\\[4pt]
  \hspace{0.8em}$\mathbf{B} \in \mathbb{R}^{D_y \times D_y}$: output coregionalization matrix
};

\end{tikzpicture}%
}

\ifdefined\GPLFRPGMINMAIN
  \GPLFRPGM
\else
  \documentclass[tikz,border=10pt]{standalone}
  \usepackage{tikz, amsmath, amssymb}
  \usetikzlibrary{positioning,arrows.meta,fit}
  \begin{document}
  \GPLFRPGM
  \end{document}
\fi\endgroup}
  \caption{Probabilistic graphical model of GPLFR. Shaded nodes are observed.}
  \label{fig:gplfr-pgm}
\end{figure*}

Although GPLFR's practical motivation is to provide an end-to-end alternative to compress-then-predict pipelines (principally PCA-GP), its mathematical structure is most transparently understood through its relationship with the LMC.
We develop these connections in Sections~\ref{sec:lmc} and \ref{sec:practice}, then compare GPLFR to PCA-GP and other baselines on a synthetic benchmark and two scientific emulation tasks: one on biomedical optics and one on the exoplanet climate problem that originally motivated this work (Section~\ref{sec:experiments}). Code is available at \url{https://github.com/edstevenson/GPLFR}.

\section{GPLFR and the Linear Model of Coregionalization (LMC)}\label{sec:lmc}

We use the LMC as a unifying lens because GPLFR and LMC can be derived from the same linear-Gaussian latent-factor model, but correspond to different marginalizations.
In particular, marginalizing the latent factors yields an LMC prior over outputs, whereas marginalizing the decoder weights yields GPLFR's collapsed likelihood used for joint representation learning and regression.
This perspective also clarifies a modeling choice emphasized in the introduction: GPLFR typically keeps any explicit output coupling (i.e., any output coupling on top of that induced by the latent factors) simple and instead concentrates modeling capacity on learning a useful similarity structure over $\mathbf{x}$.

Figure~\ref{fig:gplfr-pgm} shows the probabilistic graphical model underlying GPLFR.
A full model specification is given in Appendix~\ref{app:model}.

\paragraph{Notation.}
We consider regression from $D_x$-dimensional inputs to $D_y$-dimensional structured outputs with $N$ training examples.
Let $\mathbf{X} \in \mathbb{R}^{N \times D_x}$ denote the training inputs and $\mathbf{Y} \in \mathbb{R}^{N \times D_y}$ the corresponding outputs.
GPLFR introduces $D_z \ll D_y$ latent variables per example, collected as $\mathbf{Z} \in \mathbb{R}^{N \times D_z}$, and a linear decoder $\mathbf{W} \in \mathbb{R}^{D_y \times D_z}$.
We write $\mathbf{K}(\mathbf{X}, \mathbf{X}) \in \mathbb{R}^{N \times N}$ for kernel matrices and $\sigma^2$ for the observation noise variance.

\subsection{The LMC}\label{sec:lmc:def}

The LMC is a very general multi-output GP class that factorizes the output covariance as a sum of Kronecker products, each pairing an input kernel $\mathbf{K}_q\in \mathbb{R}^{N\times N}$ with a \emph{coregionalization} matrix $\mathbf{B}_q \in \mathbb{R}^{D_y\times D_y}$:
\begin{equation}\label{eq:lmc-cov}
  \mathrm{Cov}(\mathrm{vec}(\mathbf{Y}))
  = \sum_{q=1}^Q \mathbf{B}_q \otimes \mathbf{K}_q
  + \sigma^2 \mathbf{I}_{N D_y},
\end{equation}
where $\mathrm{vec}(\cdot)$ stacks columns.
One way to derive this is from a linear mixing of latent GP functions: for each component $q$, draw $D_q$ independent latent functions from a shared kernel $k_q$ and mix them into outputs with a matrix $\mathbf{A}_q\in \mathbb{R}^{D_y\times D_q}$.
Then $\mathbf{B}_q=\mathbf{A}_q \mathbf{A}_q^\top\succeq0$ and has rank $\leq D_q$.
The \emph{intrinsic coregionalized model} (ICM) is the special case where $\mathbf{K}_q=\mathbf{K}$ for all $q$:
\begin{equation}\label{eq:icm-cov}
  \begin{split}
    \mathrm{Cov}(\mathrm{vec}(\mathbf{Y}))
    &=\mathbf{B}\otimes \mathbf{K} + \sigma^2 \mathbf{I}_{ND_y}.
  \end{split}
\end{equation}

\subsection{GPLFR and LMC as Two Marginalizations of the Same Joint Model}\label{sec:lmc:marginalizations}

We can see the relationship between GPLFR and LMC by starting with their shared assumed data-generating process.
Partition the latent space into $Q$ groups with dimensionalities $\{D_q\}_{q=1}^Q$ such that $\sum_q D_q=D_z$.
Let $\mathbf{Z}_q\in \mathbb{R}^{N\times D_q}$ and $\mathbf{W}_q\in \mathbb{R}^{D_y\times D_q}$, and define $\mathbf{Z}=
\begin{bmatrix} \mathbf{Z}_1 & \dots & \mathbf{Z}_Q
\end{bmatrix}$ and $\mathbf{W}=
\begin{bmatrix} \mathbf{W}_1 & \dots & \mathbf{W}_Q
\end{bmatrix}$.
The data-generating process draws the latent components from independent GP priors over $\mathbf{X}$ and then maps them through a linear-Gaussian decoder:
\begin{align*}
&\textbf{Latent GP priors:}\quad \mathrm{vec}(\mathbf{Z}_q)\mid\mathbf{X}\sim \mathcal{N}\!\left(\mathbf{0},\mathbf{I}_{D_q}\otimes \mathbf{K}_q
\right)
\\& \qquad\qquad\qquad\qquad\;\;\; \text{for }q=1,\dots,Q, \\
&\textbf{Decoder:}\quad \mathbf{Y} = \sum_{q=1}^Q \mathbf{Z}_q \mathbf{W}_q^\top + \mathbf{E},\quad E_{ij}\sim \mathcal{N}(0,\sigma^2).
\end{align*}

If we marginalize out $\mathbf{Z}$, then we get
\begin{align*}
  &p(\mathrm{vec}(\mathbf{Y}) \mid \{\mathbf{W}_q\}_q, \sigma, \mathbf{X})\\
  &=\int p(\mathrm{vec}(\mathbf{Y})\mid \{\mathbf{Z}_q\}_q, \{\mathbf{W}_q\}_q, \sigma) \prod_q p(\mathbf{Z}_q \mid \mathbf{X})\,d \mathbf{Z} \\
  &=\mathcal{N}\!\left(\mathrm{vec}(\mathbf{Y}); 0, \mathbf{C}_\text{LMC}\right),
\end{align*}
with
\begin{equation}
  \mathbf{C}_\text{LMC}
  = \left[\sum_q (\mathbf{W}_q \mathbf{W}_q^\top) \otimes \mathbf{K}_q\right]
  + \sigma^2\mathbf{I}_{ND_y}.
\end{equation}
This is exactly an LMC GP~\eqref{eq:lmc-cov} with $\mathbf{B}_q=\mathbf{W}_q \mathbf{W}_q^\top$.

In GPLFR we instead marginalize out $\mathbf{W}$.
To do this we set independent matrix-normal priors on the component decoders: $\mathbf{W}_q\sim \mathcal{MN}(0, \mathbf{B}, \mathbf{I}_{D_q})$ for each $q$.
Then
\begin{align*}
  &p(\mathrm{vec}(\mathbf{Y}) \mid \{\mathbf{Z}_q\}_q, \sigma)\\
  &= \int p(\mathrm{vec}(\mathbf{Y}) \mid \{\mathbf{Z}_q\}_q, \{\mathbf{W}_q\}_q, \sigma)\,
  \prod_q p(\mathbf{W}_q)\,d \mathbf{W} \\
  &= \mathcal{N}\!\left(\mathrm{vec}(\mathbf{Y}); 0, \mathbf{C}\right),
\end{align*}
with
\begin{equation}
  \mathbf{C}
  = \mathbf{B}\otimes \left[\sum_q \mathbf{Z}_q \mathbf{Z}_q^\top \right]
  + \sigma^2 \mathbf{I}_{N D_y}.
\end{equation}
\footnote{This can also be interpreted as an ICM covariance~\eqref{eq:icm-cov} conditional on $\mathbf{Z}$, where $\mathbf{B}$ is the coregionalization matrix and the effective input-side kernel under point-estimation is $\sum_q \mathbf{Z}_q \mathbf{Z}_q^\top$.}

With this perspective, GPLFR and LMC are seen as different marginalizations of the same underlying factorization.
This is a similar idea to the \emph{primal} and \emph{dual} views of probabilistic PCA used to motivate GPLVMs in \cite{lawrenceProbabilisticNonlinear2005}.
However, note that while the primal and dual views are \emph{equivalent} (they recover the same marginal model) in the case of probabilistic PCA, the regression context of GPLFR/LMC ($\mathbf{Z}$ tied to $\mathbf{X}$ by a GP prior) immediately separates them into different model classes.\footnote{Excepting degenerate cases, e.g., if we restrict GPLFR's latent representation to be deterministic feature maps of the inputs $\mathbf{Z}_q=\boldsymbol{\Phi}_q(\mathbf{X})$ then we get ordinary LMC priors with input kernels $k_q(\mathbf{x}, \mathbf{x}'; \boldsymbol{\Phi})=\boldsymbol{\phi}_q(\mathbf{x})^\top \mathbf{T}_q \boldsymbol{\phi}_q(\mathbf{x}')$.}

\section{GPLFR: Regularization and Connection to PCA-GP}\label{sec:practice}

\subsection{Output Coregionalization and Likelihood Tempering}\label{sec:tempering}

In the high-dimensional, low-data regime that GPLFR targets, estimating a rich coregionalization matrix $\mathbf{B}$ is statistically unreliable.
We are therefore restricted to simple parameterizations of $\mathbf{B}$, relying on the latent geometry to capture most output correlations.
In practice this means setting $\mathbf{B}=\mathbf{I}$ (as we do throughout this paper), unless the output structure admits a clear low-dimensional parameterization (as in the exoplanet climate experiment).
Any such simplification is a form of model misspecification when the true output correlations contain structured variation not already captured by the latent geometry or $\mathbf{B}$'s low-dimensional parameterization.
To prevent the likelihood from overstating the information content of each output dimension, we temper it with an inverse-temperature $\beta\in (0,1]$; details and justification in Appendix~\ref{app:model:map}.

\subsection{GPLFR and PCA-GP}\label{sec:pcagp-connection}

From an application perspective, GPLFR is most naturally compared to PCA-GP -- the standard compress-then-predict pipeline for high-dimensional outputs with limited data (recapped in Appendix~\ref{app:pcagp}).
Both methods assume the outputs admit an accurate low-rank representation and fit input-to-latent mappings with GP priors, but they differ in whether that representation is learned \emph{independently of} the inputs or \emph{jointly with} the regression task.

PCA chooses a basis to retain the maximal output covariance $\mathrm{Cov}(\mathbf{y})$, essentially treating all variance as signal.
However, for prediction, we want the basis that maximally captures the \emph{predictable} component of covariance $\mathrm{Cov}(\mathbb{E}[\mathbf{y} \mid \mathbf{x}])$.
The total covariance decomposes as:
\begin{equation*}
  \begin{split}
    \mathrm{Cov}(\mathbf{y})
    &= \underbrace{\mathrm{Cov}(\mathbb{E}[\mathbf{y}\mid \mathbf{x}])}_{\text{predictable}}\; + \; \underbrace{\mathbb{E}[\mathrm{Cov}(\mathbf{y} \mid \mathbf{x})]}_{\text{unpredictable}}.
  \end{split}
\end{equation*}
When the unpredictable term is structureless (isotropic white noise), PCA is asymptotically robust as the noise merely inflates the eigenvalues uniformly.
However, when the unpredictable variation is \emph{structured} (correlated across output dimensions), it concentrates variance in specific directions independent of $\mathbf{x}$.
In this regime, PCA can misallocate capacity to high-variance but low-predictability directions.

By contrast, GPLFR's latent representation is tied to $\mathbf{x}$ by GP priors, so directions that are not coherent with $\mathbf{x}$ are penalized, biasing the representation toward predictability.
Additionally, by marginalizing decoder weights analytically, GPLFR avoids committing to a single point-estimated basis, which may be beneficial at small $N$ where a PCA basis is poorly determined.

The trade-off is that this joint objective creates a more challenging optimization landscape.
The difficulty arises chiefly from the coupling between latents and kernel hyperparameters: the kernel hyperparameters must describe the latent structure, yet that structure is itself inferred conditional on the current kernels.
We mitigate this in practice with two regularizers: a small latent noise $\lambda$ added to each latent GP covariance, which relaxes the requirement that the learned latent scores be explained exactly by smooth GP
functions of the inputs; and the likelihood tempering $\beta$ introduced above (details in Appendix~\ref{app:model:map}).
We explore the PCA-GP--GPLFR trade-off and how it depends on the characteristics of the data in Section~\ref{sec:synthetic}.

\section{Experiments}\label{sec:experiments}

We compare GPLFR to baselines on three problems: a synthetic benchmark exploring the PCA-GP--GPLFR trade-off under structured output noise (Section~\ref{sec:synthetic}); a biomedical optics emulation task in a PCA-friendly regime (Section~\ref{sec:pyxopto}); and a challenging exoplanet climate emulation task (Section~\ref{sec:exoplanet}).

\subsection{Synthetic Benchmark}\label{sec:synthetic}

\subsubsection{Set-up}\label{sec:synthetic-setup}

\paragraph{Motivation.}
This synthetic benchmark is designed to capture a common structure in real high-dimensional regression problems, where outputs consist of a low-dimensional ``signal'' component which is predictable from inputs, and residual ``nuisance'' variation that is \emph{not} predictable from inputs -- separating these two sources of variation is a key challenge for any model.
We model the nuisance component as a random field correlated across output dimensions.
This is a good stand-in for many real settings -- e.g., images, spatial fields, spectra -- where measurement effects and unresolved variability induce correlated residuals across output dimensions.
Such correlated residual structure can expose PCA-GP's failure mode of allocating capacity to high-variance but low-predictability directions.
Through the GP prior over latents, GPLFR can instead bias the learned representation toward coherence in input space (where coherence is defined by the choice of kernel family), naturally prioritizing predictable structure.

\paragraph{Data generation.}
We generate outputs $\mathbf{y} \in \mathbb{R}^{D_y}$ from inputs $\mathbf{x} \in \mathbb{R}^{D_x}$ as
\begin{equation*}
  \mathbf{y}=\mathbf{W}_\text{sig} \mathbf{z}_\text{sig}(\mathbf{x})+ \mathbf{y}_\text{nuis}+ \boldsymbol{\epsilon},
  \quad
  \boldsymbol{\epsilon}\sim \mathcal{N}(0, \sigma_\epsilon^2 \mathbf{I}_{D_y}).
\end{equation*}
Outputs live on a 2D grid with $D_y = H W$ locations.
The signal component consists of $\mathbf{z}_{\text{sig}}(\mathbf{x})\in\mathbb{R}^{D_\text{sig}}$, decoded through localized squared-exponential basis functions (columns of $\mathbf{W}_\text{sig}$) with randomly chosen centers and scales.
The nuisance component $\mathbf{y}_{\text{nuis}}\sim\mathcal{N}(0,\boldsymbol{\Sigma}_{\text{nuis}})$ is a spatially correlated random field independent of $\mathbf{x}$, alongside white noise $\boldsymbol{\epsilon}$.
Full details are in Appendix~\ref{app:synthetic:dgp}.

\paragraph{Metrics.}
Since we know the data-generating process, we can evaluate predictions against the true conditional mean $\mathbf{y}_{\text{sig}}(\mathbf{x}) \equiv \mathbb{E}[\mathbf{y}\mid\mathbf{x}] = \mathbf{W}_{\text{sig}}\,\mathbf{z}_{\text{sig}}(\mathbf{x})$, isolating signal recovery from nuisance variation.
We report
\begin{equation}
  \mathrm{RMSE}_{\text{sig}} =
  \sqrt{\frac{1}{N_{\text{test}}\, D_y} \sum_{i=1}^{N_{\text{test}}} \|\hat{\mathbf{y}}_i - \mathbf{y}_{\text{sig},i}\|_2^2}.
\end{equation}
An oracle with access to the true signal would achieve $\mathrm{RMSE}_{\text{sig}}=0$.

\paragraph{Models.}
We compare GPLFR against PCA-GP and a training-set-mean baseline.
Both models use stationary ARD RBF kernels with per-latent lengthscales.
For GPLFR, we fit the model by MAP estimation using Adam.
We set the coregionalization matrix to the identity $\mathbf{B}=\mathbf{I}$, so all output covariance modeling enters via the shared latents.
For PCA-GP, we fit the independent per-score GPs by maximizing the marginal likelihood using L-BFGS-B.
Details are in Appendix~\ref{app:synthetic:hyper}.

\subsubsection{Results}\label{sec:synthetic-results}

\begin{figure}[t]
  \vspace{-3pt}
  \centering
  \includegraphics[width=0.97\linewidth]{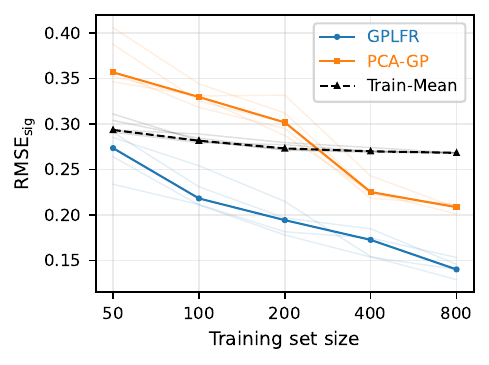}
  \caption{\emph{Synthetic benchmark:} Learning curves for GPLFR and PCA-GP, each with six latent dimensions / principal components (matching the true signal rank).
  Bold lines show medians over five dataset seeds; faint lines show individual seeds.}
  \vspace{-4pt}
  \label{fig:synthetic-learning-curves}
\end{figure}

\begin{figure}[t]
  \vspace{-3pt}
  \centering
  \includegraphics[width=0.97\linewidth]{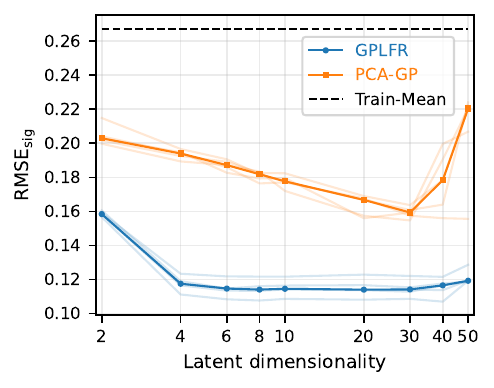}
  \caption{\emph{Synthetic benchmark:} Effect of latent dimensionality / number of principal components on signal prediction with $N=800$ examples.
    The true signal rank $D_\text{sig}=6$.
    Bold lines show medians over five dataset seeds; faint lines show individual seeds.}
  \label{fig:synthetic-latent-dim}
\end{figure}

\begin{figure}[t]
  \vspace{-3pt}
  \centering
  \includegraphics[width=1.0\linewidth]{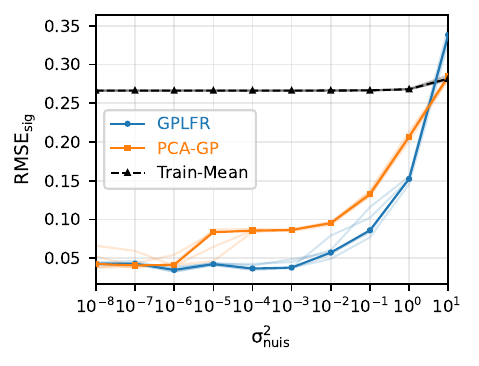}
  \caption{\emph{Synthetic benchmark:} Effect on model performance of increasing the amount of unpredictable, spatially-correlated ``nuisance'' variation $\sigma^2_\text{nuis}$ in the datasets.
  $(\sigma^2_\text{sig}, \sigma^2_\epsilon)=(1,0)$ for all datasets.
  Both models have six latent dimensions and are trained with $N=800$ examples.
  Bold lines show medians over five dataset seeds; faint lines show individual seeds.}
  \label{fig:synthetic-nuisance}
\end{figure}

For Figures~\ref{fig:synthetic-learning-curves} and \ref{fig:synthetic-latent-dim}, we use the data-generation settings: $D_y=16^2=256$, $D_x=3$, $D_\text{sig}=6$, and variances $(\sigma_\text{sig}^2, \sigma_\text{nuis}^2, \sigma_\epsilon^2)=(1,1,10^{-4})$, making a hard-to-predict dataset.

\paragraph{Sample efficiency (Figure~\ref{fig:synthetic-learning-curves}).}
Both models' RMSEs decrease roughly logarithmically with training set size $N$.
GPLFR outperforms PCA-GP across all $N$, with an advantage equivalent to roughly $4\times$ the data efficiency.

\paragraph{Representation efficiency (Figure~\ref{fig:synthetic-latent-dim}).}
GPLFR reaches its best performance (approximately) at $D_z=6$, matching the true signal rank, before plateauing and exhibiting mild overfitting at very high $D_z$.
PCA-GP improves more slowly with $D_z$ and never matches GPLFR's performance; it hits its lowest error at $D_z\approx 30$, and tends to overfit at higher $D_z$.
These discrepancies are consistent with GPLFR more effectively prioritizing signal in its latent representation, presenting the GP regressors with higher signal-to-noise regression tasks.

Since the true signal is known for this synthetic problem, we can actually directly test how much signal each model's learned subspace contains, and how much nuisance.
We project the test signal onto each model's output subspace (PCA basis for PCA-GP; decoder posterior-mean column-space for GPLFR) and measure the fraction of energy retained, and analogously for the nuisance component (details in Appendix~\ref{app:synthetic:subspace}). 
At the true signal rank $D_z=6$, GPLFR captures $77\%$ of signal energy versus $54\%$ for PCA-GP, while absorbing less nuisance ($25\%$ versus $38\%$). Figure~\ref{fig:synthetic-subspace-recovery} shows the full dependence on $D_z$.

\subsubsection{The Effect of Varying the Amount of Nuisance Variation}\label{sec:synthetic-nuisance-sweep}

The results in Figures~\ref{fig:synthetic-learning-curves} and \ref{fig:synthetic-latent-dim} show clearly the failure mode of PCA-GP and benefit of GPLFR with nuisance-heavy data.
However, the coupled GPLFR model is harder to optimize than the decoupled PCA-GP model.
This suggests a practical boundary: at what point does the flexibility of GPLFR outweigh the stability of PCA-GP?
The answer will clearly depend on the particular features of the dataset -- the kind of nuisance and signal variation, how well kernel assumptions can bias the model toward capturing signal variation etc. -- but as a first pass we can consider the effect of changing the nuisance variance across several instances of the synthetic problem.
Specifically, in Figure~\ref{fig:synthetic-nuisance}, we vary $\sigma^2_\text{nuis}$ while fixing $(\sigma^2_\text{sig}, \sigma^2_\epsilon)=(1,0)$; all other data-generating settings (including randomly generated ones) are the same as in Figures~\ref{fig:synthetic-learning-curves} and \ref{fig:synthetic-latent-dim}.

GPLFR significantly outperforms PCA-GP for $\sigma^2_\text{nuis}$ between around $10^{-5}$ and 1. We expect a large fraction of real-world problems would fall in this signal-to-nuisance ratio range.
The models converge around $\sigma^2_\text{nuis}=10^{-6}$ and are then roughly equal down to $10^{-8}$.
In the noiseless limit $\sigma^2_\text{nuis}=0$, where PCA's assumption that all variance is relevant to prediction is exactly correct, PCA-GP slightly outperforms (standard) GPLFR (we discuss this further in Appendix~\ref{app:synthetic:noiseless}).
When nuisance overwhelms signal at $\sigma^2_\text{nuis}=10$, both models overfit, underperforming the training mean baseline.

\subsection{Biomedical Optics: Emulating PyXOpto}\label{sec:pyxopto}

\subsubsection{Set-up}\label{sec:pyxopto-setup}

\paragraph{Problem description.} 
We evaluate GPLFR on a subset of \emph{MCDataset} \citep{burmenMCDatasetPublic2022}. This is a dataset from biomedical optics consisting of \emph{PyXOpto} simulations of light propagation through biological tissues.
Monte Carlo light-transport models like PyXOpto underpin a lot of non-invasive optical sensing, where one tries to infer tissue properties from measured reflectance curves for diagnosis or monitoring.
The task is to predict a $D_y=500$ dimensional radial reflectance profile $\mathbf{y}$ from three optical tissue properties $(\mu_a, \mu_s', g)$.
While $g$ is physically continuous, in the dataset it is evaluated at only $S=5$ values.
We therefore treat each $g$ value as a discrete input indexed by $s=1,\dots,S$, while treating $\mathbf{x}=(\mu_a,\mu_s')$ as a continuous input.
Monte Carlo light-transport models like PyXOpto are stochastic simulators, but noise is typically close to uncorrelated across radii, making this a relatively PCA-friendly regime, and thus a test of whether GPLFR offers any benefit when PCA's implicit assumptions are roughly met.
Further details are in Appendix~\ref{app:pyxopto:dgp}.

\paragraph{Metrics.}
We report sample-wise RMSE computed in $\log_{10}$ space on the reflectance curves,
\begin{equation}
  \mathrm{RMSE}
  = \frac{1}{N_\text{test}}\sum_{i=1}^{N_\text{test}}\frac{1}{\sqrt{D_y}}
  \big\|\log_{10}\hat{\mathbf{y}}_i-\log_{10} \mathbf{y}_i\big\|_2.
\end{equation}

\paragraph{Models.}
We compare GPLFR against PCA-ICM and a PCA-MLP baseline.
To handle the discrete nature of $g$, we equip the GP models (GPLFR and PCA-ICM) with a coregionalization kernel over the task index $s$: $k\big((\mathbf{x},s),(\mathbf{x}',s')\big) = k_x(\mathbf{x},\mathbf{x}')\,B^\text{in}_{ss'}$, where $k_x(\cdot,\cdot)$ is an ARD RBF kernel on the standardized continuous inputs and $\mathbf{B}^\text{in}\in\mathbb{R}^{S\times S}$ is a learned task correlation matrix.
This is an intrinsic coregionalized model (ICM) like we defined in Section~\ref{sec:lmc:def}, but over inputs rather than outputs.
We restrict $\mathbf{B}^\text{in}$ to a correlation matrix rather than a covariance matrix for both models to improve identifiability in this low-data regime (both models performed worse with an unrestricted covariance matrix: GPLFR slightly, PCA-ICM substantially).
GPLFR is fit by MAP estimation and PCA-ICM is fit by maximizing the marginal likelihood, both using Adam.
The MLP in PCA-MLP has two hidden layers and is trained to minimize MSE on PCA scores.
We also compare to an ``oracle'' PCA-projection baseline, which reconstructs the test outputs using the \emph{true} test PCA scores (isolating reconstruction error from regression error).
This establishes a performance ceiling for our models.
Hyperparameters and architecture details are in Appendix~\ref{app:pyxopto:hyper}.

\begin{figure}[t]
  \centering
  \includegraphics[width=1.0\linewidth]{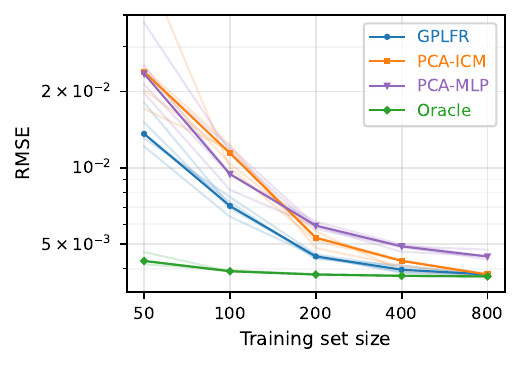}
  \caption{\emph{PyXOpto emulation:} Learning curves for GPLFR and baselines.
  GPLFR, PCA-ICM and PCA-MLP use six latent dimensions / principal components.
  Bold lines show medians over five dataset seeds; faint lines show individual seeds.}
  \label{fig:pyxopto-learning-curves}
\end{figure}
\begin{figure}[ht]
  \centering
  \includegraphics[width=1.0\linewidth]{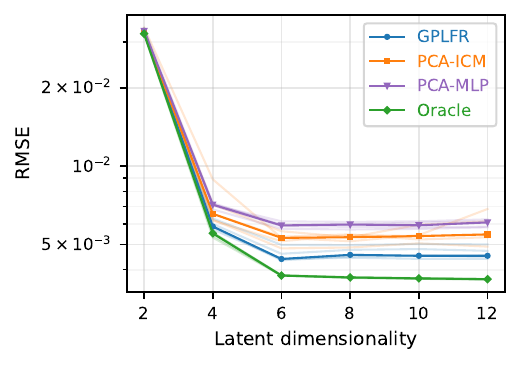}
  \caption{\emph{PyXOpto emulation:} Effect of latent dimensionality / number of principal components with $N=200$ examples.
  Bold lines show medians over five dataset seeds; faint lines show individual seeds.}
  \label{fig:pyxopto-latent-dim}
\end{figure}

\subsubsection{Results}\label{sec:pyxopto-results}
\paragraph{Sample efficiency (Figure~\ref{fig:pyxopto-learning-curves}).}
Both GP-based models converge to near the oracle benchmark at $N=800$, confirming that there is limited structured output noise in this dataset.
GPLFR shows the best sample efficiency.
\paragraph{Representation efficiency (Figure~\ref{fig:pyxopto-latent-dim}).}
With $N=200$, all models plateau around $D_z=6$, with GPLFR consistently maintaining a lower error floor.

These results suggest that even when the ``PCA assumption'' holds (signal variance $\gg$ noise variance), GPLFR can still offer greater sample and representation efficiency, due to its flexibility to rotate the representation within the signal subspace to maximize GP kernel coherence.

\subsection{Emulating Exoplanet Climate Models}\label{sec:exoplanet}

\subsubsection{Set-up}\label{sec:exoplanet-setup}

\paragraph{Scientific motivation.}
The main goal of exoplanet science is to discover alien life.
A critical subgoal is to characterize the climates of \emph{potentially habitable} exoplanets -- by which we mean planets that are Earth-like in a loose sense, spanning, for example, completely ice-covered ``snowball'' worlds to steamy ``moist greenhouse'' worlds.

The gold-standard tools for modeling these climates are global climate models (GCMs)\footnote{The acronym ``GCM'' can be read as \textbf{g}lobal \textbf{c}limate \textbf{m}odel or \textbf{g}eneral \textbf{c}irculation \textbf{m}odel; both mean the same thing.}, which numerically solve the equations of atmospheric fluid dynamics coupled to parameterizations that capture non-dynamical processes (e.g., radiation, microphysics) and sub-grid-scale dynamics (e.g., turbulence, convection).
However, a single GCM simulation typically costs $\sim 10^4$--$10^6$ core-hours, precluding their use in large parameter sweeps, uncertainty quantification, or data-driven inference.
An emulator that delivers near-instant, spatially resolved climate predictions would open up these tasks to the exoplanet community, which as of now generally must fall back to 1D models.

\begin{figure}[t]
  \centering
  \includegraphics[width=\linewidth]{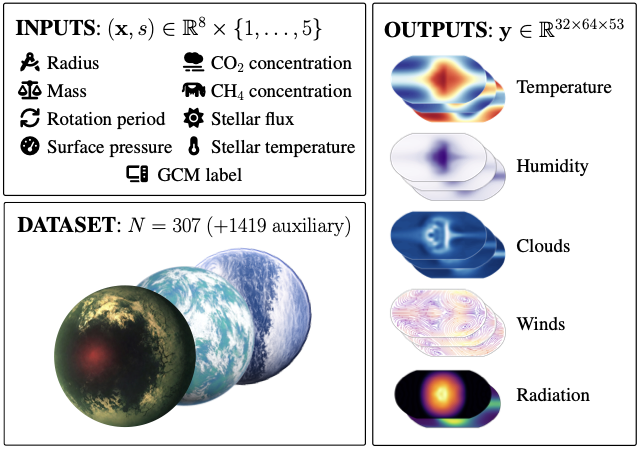}
  \caption{\emph{Exoclimate emulation:} Problem overview. Output fields are defined on a $32\times64$ latitude--longitude grid. 3D variables contain 10 pressure levels, with surface temperature recorded as a separate variable. There are 53 fields in total, giving $\sim 10^5$ output dimensions before spherical-harmonic truncation.     
  }
  \label{fig:schema}
\end{figure}

\paragraph{Problem description.}
We assemble a library of existing exoplanet GCM simulations from the literature, supplemented by bespoke runs chosen to improve input-space coverage; the combined dataset spans five GCMs and is highly non-uniform (Table~\ref{tab:exoplanet-W}).
The task is to predict the time-mean 3D atmospheric state from $D_x=8$ continuous planet properties $\mathbf{x}$ and a discrete GCM label $s\in\{1,\dots,S\}$, with $S=5$ (Figure~\ref{fig:schema}).
We use the term \emph{variable} to refer to a single physical quantity that may span multiple vertical levels while \emph{field} refers to a single slice of a variable at one vertical level. 
We represent each field on the sphere via a truncated spherical harmonic expansion and concatenate the resulting coefficients across fields, giving a single output vector $\mathbf{y}\in\mathbb{R}^{D_y}$ with $D_y\approx 3\times 10^4$.
Truncation provides a natural compact representation for smooth fields while filtering out high-frequency noise. Further preprocessing details are in Appendix~\ref{app:exoplanet:preprocessing}.

Evaluation targets two high-fidelity GCMs within a core physical domain (Table~\ref{tab:exoplanet-Y}), giving us just 315 examples, 80 of which we reserve for the test set.
To bolster the small training set, we supplement it with simulations from three auxiliary GCMs and from slightly outside the core domain, giving 1419 additional examples.

GCMs differ in their vertical grids and in which variables they output, creating structured patterns of missing data across examples.
They also embody different physics and numerics, producing systematic inter-model disagreement even for identical planets.
Even within a single GCM, differences in simulation setup (e.g., parameterization choices, time-averaging procedure, initial state) -- which are numerous, inconsistently documented, and too sparsely sampled to include as inputs -- introduce structured variability not predictable from $\mathbf{x}$ alone.

\begin{table*}[htbp]
  \centering
  \small
  \setlength{\tabcolsep}{3pt}
  \renewcommand{\arraystretch}{1.0}
  \begin{tabular}{@{}p{0.29\linewidth}rrrr *{2}{>{\raggedleft\arraybackslash}p{3.1em}}@{}} 
    \toprule
    Variable & GPLFR & PPCA-ICM & SV-LMC & PPCA-MLP & kNN & Train-mean \\
    \midrule
    Surface temperature / K & \textbf{9.32 $\pm$ 0.08} & 10.40 $\pm$ 0.02 & 10.3 $\pm$ 0.1 & 11.5 $\pm$ 0.4 & 18.3 & 28.0 \\
    Temperature (3D) / K & \textbf{7.76 $\pm$ 0.06} & 8.73 $\pm$ 0.02 & 9.66 $\pm$ 0.28 & 10.2 $\pm$ 0.3 & 14.0 & 22.2 \\
    Specific humidity (3D) / dex & \textbf{0.379 $\pm$ 0.002} & 0.402 $\pm$ 0.001 & 0.469 $\pm$ 0.008 & 0.518 $\pm$ 0.027 & 0.720 & 1.16 \\
    Cloud fraction (3D) / 1 & \textbf{0.0504 $\pm$ 0.0004} & 0.05290 $\pm$ 0.00004 & 0.0569 $\pm$ 0.0004 & 0.0591 $\pm$ 0.0007 & 0.0575 & 0.110 \\
    East--west wind (3D) / m\,s$^{-1}$ & \textbf{8.51 $\pm$ 0.07} & 9.63 $\pm$ 0.02 & 9.67 $\pm$ 0.21 & 10.4 $\pm$ 0.2 & 10.1 & 17.1 \\
    North--south wind (3D) / m\,s$^{-1}$ & \textbf{3.70 $\pm$ 0.03} & 4.31 $\pm$ 0.01 & 3.88 $\pm$ 0.02 & 4.47 $\pm$ 0.13 & 4.34 & 6.39 \\
    Absorbed shortwave radiation / W\,m$^{-2}$ & \textbf{19.6 $\pm$ 0.7} & 23.5 $\pm$ 0.8 & 22.5 $\pm$ 0.9 & 28.7 $\pm$ 0.9 & 47.7 & 343 \\
    Outgoing longwave radiation / W\,m$^{-2}$ & \textbf{12.8 $\pm$ 0.1} & 15.30 $\pm$ 0.02 & 16.2 $\pm$ 0.2 & 16.5 $\pm$ 0.3 & 28.9 & 48.2 \\
    \bottomrule
  \end{tabular}
  \caption{\emph{Exoclimate emulation:} RMSE. Learned methods are reported as mean $\pm$ standard deviation over five training seeds (kNN and the training mean are deterministic). Bold indicates the lowest (best) value. 3D variables are shown averaged over pressure levels; per-level results are reported in Table~\ref{tab:exoplanet-AA}.}
  \label{tab:exoplanet-rmse-summary}
  
\end{table*}
\begin{table*}[htbp]
  \small
  \setlength{\tabcolsep}{3pt}
  \renewcommand{\arraystretch}{1.0}
  \begin{flushleft}
  \begin{tabular}{@{}p{0.29\linewidth}rrr@{}}
    \toprule
    Variable & GPLFR & PPCA-ICM & SV-LMC \\
    \midrule
    Surface temperature / K & \textbf{7.17 $\pm$ 0.08} & 8.25 $\pm$ 0.03 & 9.28 $\pm$ 0.05 \\
    Temperature (3D) / K & \textbf{6.01 $\pm$ 0.03} & 7.03 $\pm$ 0.02 & 8.53 $\pm$ 0.13 \\
    Specific humidity (3D) / dex & \textbf{0.660 $\pm$ 0.006} & 0.723 $\pm$ 0.003 & 0.969 $\pm$ 0.006 \\
    Cloud fraction (3D) / 1 & \textbf{0.0376 $\pm$ 0.0002} & 0.04010 $\pm$ 0.00004 & 0.0413 $\pm$ 0.0003 \\
    East--west wind (3D) / m\,s$^{-1}$ & \textbf{6.65 $\pm$ 0.06} & 7.67 $\pm$ 0.02 & 7.34 $\pm$ 0.19 \\
    North--south wind (3D) / m\,s$^{-1}$ & \textbf{2.91 $\pm$ 0.02} & 3.45 $\pm$ 0.01 & 2.91 $\pm$ 0.02 \\
    Absorbed shortwave radiation / W\,m$^{-2}$ & \textbf{14.8 $\pm$ 0.7} & 18.0 $\pm$ 0.6 & 31.5 $\pm$ 0.4 \\
    Outgoing longwave radiation / W\,m$^{-2}$ & \textbf{9.76 $\pm$ 0.17} & 11.40 $\pm$ 0.03 & 15.6 $\pm$ 0.1 \\
    \bottomrule
  \end{tabular}
  \end{flushleft}
  \caption{\emph{Exoclimate emulation:} Energy scores, reported as mean $\pm$ standard deviation over five training seeds. Bold indicates the lowest (best) value. 3D variables are shown averaged over pressure levels; per-level results are reported in Table~\ref{tab:exoplanet-Z}.}
  \label{tab:exoplanet-energy-summary}
\end{table*}

\paragraph{Models.}
We compare six models: GPLFR, PPCA-ICM, and SV-LMC are probabilistic models that operate in spectral space and produce ensemble predictions for uncertainty quantification; PPCA-MLP, $k$-nearest neighbours (kNN), and the training mean serve as deterministic baselines. 

GPLFR fits independent ARD Mat\'ern-$\tfrac{5}{2}$ ICM GPs over $(\mathbf{x},s)$ with an identity decoder ($\mathbf{B}=\mathbf{I}$). Missing fields are masked in the collapsed likelihood. Details in Appendix~\ref{app:exoplanet:gplfr}.

PPCA-ICM extracts latent scores with probabilistic PCA, then regresses them against $(\mathbf{x},s)$ using independent ICM GPs as in GPLFR. Missing fields are omitted from the PPCA likelihood. Details in Appendix~\ref{app:exoplanet:ppca}.

SV-LMC represents LMC-style models (Section~\ref{sec:lmc:def}), made tractable here by sparse variational GP (SV) inference \citep{hensmanGaussianProcesses2013,nguyenCollaborativeMultioutput2014,wilkFrameworkInterdomain2020}: $D_z$ latent GPs, each with its own ARD Mat\'ern-$\tfrac{5}{2}$ kernel over $\mathbf{x}$ multiplied by a coregionalization factor over $s$ shared across latents, are mixed into the outputs by a learned unrestricted $D_y\times D_z$ loading matrix. Missing fields are dropped from the likelihood. Details in Appendix~\ref{app:exoplanet:svlmc}.

PPCA-MLP replaces PPCA-ICM's ICM GP with a two-layer MLP. kNN and the training mean operate directly in spatial space. Details in Appendix~\ref{app:exoplanet:baselines}.

\paragraph{Metrics.}
We evaluate predictions on spatial fields after full inverse preprocessing.
For deterministic evaluation, we report the area-weighted RMSE:
\begin{equation*}
  \mathrm{RMSE}(\hat{\mathbf{y}},\mathbf{y})=\|\hat{\mathbf{y}}-\mathbf{y}\|_G,
  \quad
  \|\mathbf{e}\|_G=\sqrt{\mathbf{e}^\top\mathbf{G}\mathbf{e}},
\end{equation*}
where $\mathbf{G}$ contains Gauss--Legendre latitude weights, and $\mathbf{y}$ and $\hat{\mathbf{y}}$ are the true and predicted fields.
For probabilistic evaluation, we report the energy score\footnote{The energy score can be seen as a multivariate generalization of the continuous ranked probability score (CRPS). The first term penalizes mean inaccuracy and the second penalizes overconfidence; lower is better.} of the predictive distribution $p$ against a ground-truth field $\mathbf{y}$
\begin{equation*}
  \mathrm{ES}(p,\mathbf{y})=
  \mathbb{E}_{\mathbf{y}'\sim p}\|\mathbf{y}'-\mathbf{y}\|_G
  -\tfrac{1}{2}\mathbb{E}_{\mathbf{y}',\mathbf{y}''\sim p}\|\mathbf{y}'-\mathbf{y}''\|_G,
\end{equation*}
estimated using $64$ posterior predictive samples (Appendix~\ref{app:exoplanet:energy-score-estimator}).
Both metrics are computed per-field and averaged over test examples.

\subsubsection{Results}\label{sec:exoplanet-results}
\vspace{-0pt}
GPLFR achieves the lowest RMSE and energy score across all variables (Tables~\ref{tab:exoplanet-rmse-summary} and \ref{tab:exoplanet-energy-summary}), with RMSE improvements over the next best model ranging from 5\% (north--south wind) to 16\% (outgoing longwave radiation); energy score improvements are similar (although GPLFR is only marginally ahead on north--south wind). PPCA-ICM is the next-best model overall, although SV-LMC is slightly stronger on some variables.
Appendix~\ref{app:exoplanet:uncertainty-scores} reports paired bootstrap intervals for the RMSE results to quantify uncertainty from the test-set draw.
Example spatial predictions are shown in Appendix~\ref{app:exoplanet:examples}.

We note the contrast with the PyXOpto experiment.
There, GPLFR and PPCA-ICM were much closer in performance, consistent with the simulator noise being nearly uncorrelated across output dimensions -- a regime where PCA's implicit assumption that all variance is relevant to prediction is approximately correct.
The large performance gap here suggests that, as in the synthetic benchmark (Section~\ref{sec:synthetic}), PPCA-ICM is allocating modeling capacity to less predictable directions than GPLFR -- plausibly due to structured, input-independent variation arising from unmodeled differences in simulation setup.

\section{Limitations and Future Work}\label{sec:limitations}

\paragraph{Optimization landscape.}
Unlike the sequential estimation in PCA-GP, GPLFR requires jointly optimizing latents and kernel hyperparameters.
This introduces a strong coupling: the kernel hyperparameters determine the prior induced on the latent space, yet the latent variables are identifiable only conditional on that kernel.
Consequently, the objective is highly multimodal.
In practice one can stabilize fitting with likelihood tempering $\beta$ and latent noise $\lambda$ (Section~\ref{sec:practice}), but these do increase the hyperparameter tuning burden; better heuristics for choosing these parameters would reduce this cost.

\paragraph{Structured residuals.}
Our current implementation defaults to an identity coregionalization matrix ($\mathbf{B}=\mathbf{I}$), relying on likelihood tempering to mitigate the resulting misspecification when residuals are correlated.
Future work could explore low-dimensional parameterizations of $\mathbf{B}$, such as Kronecker products for gridded data or Toeplitz matrices for time-series.

\paragraph{Inference and uncertainty quantification.}
Here we focus on fitting GPLFR via MAP estimation.
Distributional approximations to the full posterior, such as variational inference or Hamiltonian Monte Carlo (HMC), are an avenue for future work, and may be particularly beneficial in very low data (or high compute) settings.
A promising middle ground is a ``partially Bayesian'' approximation: fixing the latents at their MAP estimates while performing Bayesian inference (e.g., via HMC) over the remaining parameters, avoiding the cost and identifiability challenge of sampling many latents.
This is reasonable in the low-data, high-dimensional outputs regime because the posterior over latents is typically much more concentrated, given many output dimensions per example, than the posterior over the remaining parameters, which are only weakly constrained given few examples.

\section{Conclusion}\label{sec:conclusion}

We have presented GPLFR, a probabilistic model that couples representation learning and regression by learning latents under a GP prior jointly with output reconstruction through a collapsed linear-Gaussian decoder.
This biases the learned representation towards structure that is \emph{predictable} from the inputs, which is especially useful in problems with structured output noise.
On the synthetic benchmark, GPLFR achieves the same error as PCA-GP with roughly $4\times$ fewer training data; on PyXOpto emulation, it shows consistent sample efficiency gains even in a PCA-friendly regime with limited structured output noise.
The main practical cost relative to PCA-GP is harder optimization, which can be mitigated with regularization.
Beyond these benchmarks, we used GPLFR to build the first spatially resolved emulator of global climate models for rocky exoplanets, where it strongly outperformed alternative methods.

\begin{acknowledgements}
  Edward Stevenson is supported by the Science and Technology Facilities Council (STFC) Centre for Doctoral Training in Data Intensive Science at the University of Cambridge.
  Miles Cranmer is grateful for support from the Isaac Newton Trust and the AI2050 program at Schmidt Sciences.
  Mei Ting Mak acknowledges support from the Croucher Postdoctoral Fellowship, funded by the Croucher Foundation.
  The GCM results are produced using Met Office Software and the Monsoon3 system, a collaborative facility supplied under the Joint Weather and Climate Research Programme, a strategic partnership between the Met Office and the Natural Environment Research Council in the UK.
  Eric Wolf acknowledges funding from the Consortium on Habitability and Atmospheres of M-dwarf Planets team and the Virtual Planetary Laboratory, supported by NASA grant numbers 80NSSC21K0905, 80NSSC23K1399, 80NSSC23K1398 and 80NSSC18K0829 respectively.
  Nathan Mayne acknowledges support from a UK Research and Innovation (UKRI) Future Leaders Fellowship MR/T040866/1, and partly from the Leverhulme Trust through a research project grant RPG-2020-82 alongside a Science and Technology Facilities Council (STFC) Consolidated Grant ST/R000395/1.
  This work used the Dawn AI service, part of the UK AI Research Resource (AIRR), operated by the University of Cambridge Research Computing Service (\url{https://www.hpc.cam.ac.uk/d-w-n}) and supported by UK Research and Innovation, with Intel and Dell Technologies as technology partners.
\end{acknowledgements}

% \newpage

% References
\bibliography{My_Library,manual}

\newpage
\onecolumn

\title{Gaussian Process Latent Factor Regression for Low-Data, High-Dimensional Output Problems\\(Supplementary Material)}
\maketitle
% Preprint: add \thispagestyle{numbered} here if the first appendix page loses its number; remove with the other preprint page-style lines for UAI camera-ready.

\appendix
\counterwithin{figure}{section}
\counterwithin{table}{section}
\renewcommand{\thefigure}{\Alph{section}.\arabic{figure}}
\renewcommand{\thetable}{\Alph{section}.\arabic{table}}

\section{Adjacent Task-Aware Representation Learning Methods}\label{app:adjacent-task-aware}

Compress-then-predict pipelines learn a basis optimized for output reconstruction, not predictability from the inputs -- a well-known limitation. Several alternatives exist, but none are satisfactory in our setting.
Supervised PCA \citep{bairPredictionSupervised2006} screens output dimensions by their relevance to the inputs before compression; however, when nuisance variation is distributed across output dimensions rather than confined to a subset -- as is typical of structured outputs like spatial fields -- screening individual dimensions cannot separate signal from nuisance.
What is needed is a change of the basis itself, informed by the inputs.
Partial least squares \citep{woldPLSregressionBasic2001} and reduced-rank regression \citep{izenmanReducedrankRegression1975} achieve this, but assume a linear relationship between inputs and outputs.
On the input side, Manifold GPs \citep{calandraManifoldGaussian2016} jointly learn a nonlinear input transformation with GP regression, demonstrating the benefit of task-aware representation learning -- but for scalar or low-dimensional outputs, not high-dimensional output compression.
End-to-end neural approaches learn nonlinear input-aware representations of both inputs and outputs but are typically data-hungry and lack principled uncertainty quantification.

\section{GPLFR: Model Specification and Inference}\label{app:model}

This section complements Section~\ref{sec:lmc} by writing the GPLFR probabilistic model explicitly and stating the objective used for fitting and prediction.
We present the most flexible special case: per-latent kernels with separate lengthscales and amplitudes (i.e., $D_q=1\ \forall q$); restrictions to various lengthscale or amplitude groupings are straightforward to implement and can be useful in weakly identified settings.

\subsection{Notation and Data Layout}\label{app:model:notation}

\paragraph{Inputs.}
An input is $\mathbf{x}\in \mathbb{R}^{D_x}$, with training inputs $\mathbf{X}=
\begin{bmatrix} \mathbf{x}_1^\top \\ \vdots \\ \mathbf{x}_N^\top
\end{bmatrix}\in\mathbb{R}^{N\times D_x}$.

\paragraph{Outputs.}
An output is $\mathbf{y}\in \mathbb{R}^{D_y}$, with training outputs $\mathbf{Y}=
\begin{bmatrix} \mathbf{y}_1^\top \\ \vdots \\ \mathbf{y}_N^\top
\end{bmatrix}\in\mathbb{R}^{N\times D_y}$.
We assume centered outputs.

\paragraph{Latents and decoder.}
Latent variables for the training set are collected in $\mathbf{Z}\in\mathbb{R}^{N\times D_z}$ and decoder weights in $\mathbf{W}\in\mathbb{R}^{D_y\times D_z}$.
We view both as concatenations of $Q$ blocks: $\mathbf{Z}=
\begin{bmatrix} \mathbf{Z}_1 & \dots & \mathbf{Z}_Q
\end{bmatrix}$ and $\mathbf{W}=
\begin{bmatrix} \mathbf{W}_1 & \dots & \mathbf{W}_Q
\end{bmatrix}$, where $\mathbf{Z}_q\in\mathbb{R}^{N\times D_q}$, $\mathbf{W}_q\in\mathbb{R}^{D_y\times D_q}$, and $\sum_q D_q=D_z$.
The per-latent special case used in this paper has $D_q=1$ for all $q$, so we index by $q=1,\dots,D_z$ and let $\mathbf{z}^{(q)}\in\mathbb{R}^{N}$ and $\mathbf{w}^{(q)}\in\mathbb{R}^{D_y}$ denote the $q$-th latent variable (evaluated at the training inputs) and its corresponding decoder column.\footnote{The alternative marginalization (as per Section~\ref{sec:lmc:marginalizations}) for the per-latent case yields the semiparametric latent factor model (SLFM) of \citet{tehSemiparametricLatent2005}, with $\mathbf{C}_\text{LMC} =\sum_{q=1}^{D_z} (\mathbf{w}^{(q)} \mathbf{w}^{(q)\top}) \otimes \mathbf{K}_q+ \sigma^2 \mathbf{I}_{ND_y}$.}

\paragraph{Parameters and hyperparameters.}
We refer to quantities inferred during model fitting as \emph{parameters}, and fixed design choices made during model selection as \emph{hyperparameters} (omitted from conditioning notation).
Within \emph{parameters}, we distinguish between the latents $\mathbf{Z}$, and the remaining ``global'' parameters.

\subsection{Generative Model}\label{app:model:generative}

GPLFR couples a GP ``encoder'' prior over latents with a linear-Gaussian ``decoder'' from latents to outputs.

\paragraph{Encoder.}
For each latent dimension $q\in\{1,\dots,D_z\}$ we introduce a latent function $z^{(q)}(\cdot):\mathbb{R}^{D_x}\to\mathbb{R}$ with an independent zero-mean GP prior $z^{(q)}(\cdot)\sim\mathcal{GP}(0,k_q(\cdot, \cdot))$.
On the training inputs this implies
\begin{equation*}
  \mathbb{R}^N\ni\mathbf{z}^{(q)}\mid \mathbf{X}, \boldsymbol{\ell}_q, \eta_q \sim \mathcal{N}(0, \mathbf{K}_q),
  \quad
  K_{q,ij}=k_q(\mathbf{x}_i, \mathbf{x}_j;\boldsymbol{\ell}_q,\eta_q).
\end{equation*}
	Each kernel is parameterized by ARD lengthscales $\boldsymbol{\ell}_q\in\mathbb{R}^{D_x}$ and an amplitude $\eta_q$.
	In practice, we use a regularized kernel $\mathbf{K}_q \leftarrow \mathbf{K}_q + \lambda \mathbf{I}_N$, where the latent noise $\lambda$ is a hyperparameter, to help stabilize optimization.
	Collecting across $q$ gives
	\begin{equation*}
	  p(\mathbf{Z}\mid \mathbf{X},\{\boldsymbol{\ell}_q,\eta_q\}_q)=\prod_{q=1}^{D_z}p(\mathbf{z}^{(q)}\mid \mathbf{X},\boldsymbol{\ell}_q,\eta_q).
	\end{equation*}

\paragraph{Decoder.}
We use a linear-Gaussian decoder
\begin{equation*}
  \mathbf{y}\mid \mathbf{z},\mathbf{W},\sigma \sim \mathcal{N}(\mathbf{W}\mathbf{z},\sigma^2\mathbf{I}_{D_y})
\end{equation*}
with weight prior
\begin{equation*}
  \mathbf{W}\sim \mathcal{MN}\big(0,\mathbf{B},\mathbf{I}_{D_z}\big).
\end{equation*}
Here $\mathbf{B}\in\mathbb{R}^{D_y\times D_y}$ is an output-space row covariance.
In the high-dimensional output regime we default to $\mathbf{B}=\mathbf{I}_{D_y}$; structured choices for $\mathbf{B}$ are also possible.

\subsection{Collapsed Decoder Likelihood}\label{app:model:collapsed}

For one output dimension $j$ (a column of $\mathbf{Y}$),
\begin{equation*}
  \mathbf{y}^{(j)}\mid \mathbf{Z},\mathbf{w}^{(j)},\sigma \sim \mathcal{N}(\mathbf{Z}\mathbf{w}^{(j)},\sigma^2\mathbf{I}_N).
\end{equation*}
Marginalizing out $\mathbf{W}\sim \mathcal{MN}\big(0,\mathbf{B},\mathbf{I}_{D_z}\big)$ yields
\begin{equation*}
  \mathrm{vec}(\mathbf{Y})\mid \mathbf{Z},\mathbf{B},\sigma \sim \mathcal{N}(0,\mathbf{C}),
  \quad
  \mathbf{C}=\mathbf{B}\otimes\big(\mathbf{Z} \mathbf{Z}^\top\big)+\sigma^2\mathbf{I}_{ND_y}.
\end{equation*}
This is the prior predictive distribution of $\mathbf{Y}$ conditional on $\mathbf{Z}$.

For the independence case $\mathbf{B}=\mathbf{I}_{D_y}$ we use throughout this paper, because all columns are conditionally independent given $\mathbf{Z}$, this reduces to
\begin{equation*}
  p(\mathbf{Y}\mid \mathbf{Z},\sigma)=\prod_{j=1}^{D_y}\mathcal{N}\big(\mathbf{y}^{(j)};0,\mathbf{C}_N\big),
  \quad
  \mathbf{C}_N=\mathbf{Z}\mathbf{Z}^\top+\sigma^2\mathbf{I}_N.
\end{equation*}

\paragraph{Efficient evaluation of the collapsed likelihood.}
The collapsed likelihood requires $\mathbf{C}^{-1}$ and $\log\det\mathbf{C}$.
We evaluate these via Woodbury identities for diagonal plus low-rank structure.
For the $\mathbf{B}=\mathbf{I}_{D_y}$ case, by the matrix inversion lemma (Woodbury identity) and matrix determinant lemma,
\begin{equation*}
  \mathbf{C}_N^{-1}=\frac{1}{\sigma^2}\big(\mathbf{I}_N-\mathbf{Z}\mathbf{D}^{-1}\mathbf{Z}^\top\big),
  \quad  
  \log\det\mathbf{C}_N=(N-D_z)\log\sigma^2+\log\det\mathbf{D},
\end{equation*}
where $\mathbf{D}=\sigma^2\mathbf{I}_{D_z}+\mathbf{Z}^\top\mathbf{Z}$.
The per-column quadratic form is
\begin{equation*}
  \mathbf{y}^{(j)\top}\mathbf{C}_N^{-1}\mathbf{y}^{(j)}=\frac{1}{\sigma^2}\left(\|\mathbf{y}^{(j)}\|_2^2-(\mathbf{Z}^\top\mathbf{y}^{(j)})^\top\mathbf{D}^{-1}(\mathbf{Z}^\top\mathbf{y}^{(j)})\right).
\end{equation*}
Summing over $j$, the computational complexity is $O(N D_z D_y + D_z^2 D_y + N D_z^2+D_z^3)$ which is typically dominated by the $O(ND_zD_y)$ term.
The same approach applies for general $\mathbf{B}$ using Kronecker structure.

\subsection{Posterior and MAP Estimation}\label{app:model:map}

Let $\boldsymbol{\phi}\equiv\{\mathbf{Z},\mathbf{B},\sigma,\{\boldsymbol{\ell}_q,\eta_q\}_q\}$ collect the parameters and $\mathcal{D}\equiv\{\mathbf{X},\mathbf{Y}\}$ the dataset.
The (tempered) posterior is
\begin{equation}\label{eq:tempered-posterior}
  p(\boldsymbol{\phi}\mid\mathcal{D})\propto p(\mathbf{Y}\mid \mathbf{Z},\mathbf{B},\sigma)^\beta\,p(\mathbf{Z}\mid \mathbf{X},\{\boldsymbol{\ell}_q,\eta_q\}_q)\,p(\{\boldsymbol{\ell}_q,\eta_q\}_q)\,p(\sigma^2)\,p(\mathbf{B}).
\end{equation}
We approximate this posterior as $p(\boldsymbol{\phi}\mid\mathcal{D})\approx\delta(\boldsymbol{\phi}-\boldsymbol{\phi}^\star)$ where
\begin{equation*}
  \boldsymbol{\phi}^\star=\arg\max_{\boldsymbol{\phi}}\log p(\boldsymbol{\phi}\mid\mathcal{D}).
\end{equation*}
The priors $p(\{\boldsymbol{\ell}_q,\eta_q\}_q),p(\sigma^2),p(\mathbf{B})$ act as regularizers under MAP.

\paragraph{Likelihood tempering.}
We use a tempered likelihood with inverse-temperature $\beta\in(0,1]$ to mitigate decoder misspecification.
This misspecification is most acute when outputs exhibit strong output correlations not well approximated by the low-rank latent-factor structure; for example, with $\mathbf{B}=\mathbf{I}$, GPLFR assumes no residual covariance beyond that induced by the shared latents.
The untempered likelihood can therefore overstate how informative each output dimension is when (i) the correlated residual structure is high-rank, or (ii) correlations have a large residual component that is weakly tied to $\mathbf{x}$ (e.g.\ input-independent nuisance fields, or correlation patterns that vary across $\mathbf{x}$ in ways a global linear decoder cannot capture).
Tempering in response to model misspecification has a decision-theoretic justification as a generalized Bayesian update \citep{bissiriGeneralFramework2016}, and, practically, in GPLFR an untempered likelihood can overwhelm the GP prior, leading to a brittle latent geometry. 

\paragraph{Optimization.}
We jointly optimize the (tempered) log-posterior~\eqref{eq:tempered-posterior} via Adam, using separate learning rates for the latents $\mathbf{Z}$ (which are local to datapoints) and global parameters (in general: $\{\boldsymbol{\ell}_q, \eta_q\}_q, \sigma, \mathbf{B}$) since the effective gradient scale differs between the two groups.

\subsection{Prediction}\label{app:model:prediction}

For a test input $\mathbf{x}_*$, the posterior predictive distribution is
\begin{align*}
  p(\mathbf{y}_*\mid \mathbf{x}_*,\mathcal{D})
  &=\int p(\mathbf{y}_*\mid \mathbf{x}_*,\mathbf{Y},\boldsymbol{\phi})\,p(\boldsymbol{\phi}\mid\mathcal{D})\,d\boldsymbol{\phi}\\
  &\approx p(\mathbf{y}_*\mid \mathbf{x}_*,\mathbf{Y},\boldsymbol{\phi}^\star)\qquad\text{(MAP approximation).}
\end{align*}
We describe sampling from the MAP-approximated predictive distribution in two steps: encoder then decoder (we omit the ${}^\star$ superscript for brevity).

\paragraph{1.\ Encoder.}
Sample the test latent from the posterior predictive over latents $p(\mathbf{z}_* \mid \mathbf{Z}, \mathbf{X}, \{\boldsymbol{\ell}_q\}_q)$.
For dimension $q$,
\begin{equation*}
  z_*^{(q)}\mid \mathbf{z}^{(q)},\mathbf{X},\boldsymbol{\ell}_{q},\eta_q \sim \mathcal{N}\!\Big( \mathbf{k}_{*,q}^\top \mathbf{K}_q^{-1}\mathbf{z}^{(q)},\  k_{**,q}-\mathbf{k}_{*,q}^\top \mathbf{K}_q^{-1}\mathbf{k}_{*,q} \Big),
\end{equation*}
where
\begin{equation*}
  K_{q,ij}=k_q\big(\mathbf{x}_i,\mathbf{x}_j;\boldsymbol{\ell}_q, \eta_q\big),\quad
  \mathbf{k}_{*,q}=k_q\big(\mathbf{x}_*,\mathbf{X};\boldsymbol{\ell}_q, \eta_q\big),\quad
  k_{**,q}=k_q\big(\mathbf{x}_*, \mathbf{x}_*;\boldsymbol{\ell}_q, \eta_q\big).
\end{equation*}
Stacking these over $q$ gives
\begin{equation*}
  \mathbf{z}_*\mid \mathbf{Z} ,\mathbf{X},\{\boldsymbol{\ell}_{q}, \eta_q\}_q \sim \mathcal{N}\!\big(\boldsymbol{\mu}_{z_*},\; \mathrm{diag}(\boldsymbol{\sigma}^2_{z_*})\big).
\end{equation*}

\paragraph{2.\ Decoder.}
Sample the test output given the test latent from the latent-conditional posterior predictive $p(\mathbf{y}_* \mid \mathbf{z}_*,\mathbf{Y}, \mathbf{Z} , \mathbf{B} , \sigma )$.
After marginalizing $\mathbf{W}$ and conditioning on $\mathbf{z}_*$, the joint Gaussian over training outputs and the test output is
\begin{equation*}
  \begin{bmatrix} \mathrm{vec}(\mathbf{Y}) \\ \mathbf{y}_*
  \end{bmatrix} \Bigm| \; \mathbf{z}_*,\mathbf{Z} , \mathbf{B} , \sigma
  \sim
  \mathcal{N}\!\left( \mathbf{0},
    \begin{bmatrix} \mathbf{C}_{YY} & \mathbf{C}_{Y*} \\ \mathbf{C}_{*Y} & \mathbf{C}_{**}
  \end{bmatrix} \right),
\end{equation*}
with
\begin{equation*}
  \mathbf{C}_{YY}=\mathbf{B}\otimes (\mathbf{Z} \mathbf{Z}^\top)+\sigma^2 \mathbf{I}_{ND_y},\quad
  \mathbf{C}_{Y*}=\mathbf{B}\otimes(\mathbf{Z} \mathbf{z}_*),\quad
  \mathbf{C}_{**} = \mathbf{B}\otimes(\mathbf{z}_*^\top  \mathbf{z}_*)  + \sigma^2 \mathbf{I}_{D_y}.
\end{equation*}
Conditioning on observed $\mathbf{Y}$ gives
\begin{equation*}
  \mathbf{y}_*\mid \mathbf{z}_*,\mathbf{Y},\mathbf{Z},\mathbf{B},\sigma \sim \mathcal{N}(\boldsymbol{\mu}_{y_*},\boldsymbol{\Sigma}_{y_*}),
  \quad
  \boldsymbol{\mu}_{y_*}=\mathbf{C}_{*Y}\mathbf{C}_{YY}^{-1}\mathrm{vec}(\mathbf{Y}),
  \quad
  \boldsymbol{\Sigma}_{y_*}=\mathbf{C}_{**}-\mathbf{C}_{*Y}\mathbf{C}_{YY}^{-1}\mathbf{C}_{Y*}.
\end{equation*}
When $\mathbf{B}=\mathbf{I}_{D_y}$, $\mathbf{C}_{YY}$ is block-diagonal and this reduces to independent per-$j$ conditioning with $N\times N$ covariance $\mathbf{C}_N=\mathbf{Z}\mathbf{Z}^\top+\sigma^2\mathbf{I}_N$.

\paragraph{Predictive mean and covariance.}
With samples $\{\mathbf{y}_*^{[m]}\}_{m=1}^M$ from the MAP predictive,
\begin{align*}
  \mathbb{E}(\mathbf{y}_*\mid\mathcal{D})&\approx \hat{\mathbf{y}}_*\equiv\frac{1}{M}\sum_m \mathbf{y}_*^{[m]},\\
  \mathrm{Cov}(\mathbf{y}_*\mid\mathcal{D})&\approx \hat{\boldsymbol{\Sigma}}_*\equiv \frac{1}{M-1}\sum_m\big(\mathbf{y}_*^{[m]}-\hat{\mathbf{y}}_*\big)\big(\mathbf{y}_*^{[m]}-\hat{\mathbf{y}}_*\big)^\top.
\end{align*}
In practice, when uncertainty is not required, we compute $\hat{\mathbf{y}}_*$ using the analytic predictive means of the encoder and decoder conditionals to avoid Monte Carlo noise and reduce compute.

\section{PCA-GP}\label{app:pcagp}

Gaussian process regression on PCA scores (PCA-GP, including the extensions below) is the key baseline in our experiments, since it is the standard compress-then-predict model for low-data, high-dimensional outputs, and is structurally similar.
Here we briefly review PCA-GP and state the base configuration used in our experiments.

PCA-GP is a two-stage compress-then-predict pipeline.
First, we fit PCA on the (centered) training outputs to obtain a $D_z$-dimensional basis $\boldsymbol{\Phi} \in \mathbb{R}^{D_z \times D_y}$, with training scores $\mathbf{w}_i = \boldsymbol{\Phi}\mathbf{y}_i$.
Second, we regress each score dimension independently against the inputs with a GP.
For each $q\in\{1,\dots,D_z\}$ we model
\begin{equation*}
  w_{i,q}=f_q(\mathbf{x}_i)+\xi_{i,q},
  \quad
  f_q(\cdot)\sim\mathcal{GP}(0,k_q(\cdot, \cdot)),
  \quad
  \xi_{i,q}\sim\mathcal{N}(0,\sigma_{\xi,q}^2).
\end{equation*}
As with GPLFR, it is a modeling choice how strongly to tie kernel parameters across components, including the score noise (nugget) term.
We found a shared score noise variance to be a robust default and fix it across experiments, while the degree of sharing for kernel lengthscales and amplitudes is chosen on a per-experiment basis.
Given a test input $\mathbf{x}_*$, we form $\hat{\mathbf{w}}_*$ by GP prediction in each score dimension and predict the output as $\hat{\mathbf{y}}_*=\boldsymbol{\Phi}^\top\hat{\mathbf{w}}_*$.
Predictive uncertainty is obtained by combining the independent per-score GP predictive variances through the linear reconstruction.

\section{Experiment Details}\label{app:experiments}

\subsection{Synthetic Benchmark}\label{app:synthetic}

\subsubsection{Data Generating Process}\label{app:synthetic:dgp}

We generate outputs $\mathbf{y} \in \mathbb{R}^{D_y}$ from inputs $\mathbf{x} \in \mathbb{R}^{D_x}$ as
\begin{equation*}
  \mathbf{y}=\mathbf{W}_\text{sig} \mathbf{z}_\text{sig}(\mathbf{x})+ \mathbf{y}_\text{nuis}+ \boldsymbol{\epsilon},
  \quad
  \boldsymbol{\epsilon}\sim \mathcal{N}(0, \sigma_\epsilon^2 \mathbf{I}_{D_y}).
\end{equation*}

\paragraph{Predictable variation.}
Choose $D_\text{sig}$ signal latents and draw each as a function of $\mathbf{x}$:
\begin{equation*}
  z_\text{sig}^{(q)}(\cdot)\sim \mathcal{GP}(0,\,\sigma_\text{sig}^2 k(\cdot, \cdot ))\quad \text{for } q=1,\dots,D_\text{sig}.
\end{equation*}
For training inputs $\{\mathbf{x}_i\}_{i=1}^N$, this implies
\begin{equation*}
  \mathbf{z}^{(q)}_\text{sig}\equiv\big(z_\text{sig}^{(q)}(\mathbf{x}_1),\dots,z_\text{sig}^{(q)}(\mathbf{x}_N)\big)^\top \sim \mathcal{N}(0,\, \sigma_\text{sig}^2 \mathbf{K}),
  \quad
  K_{ij}=k(\mathbf{x}_i, \mathbf{x}_j).
\end{equation*}
Stacking across $q$ gives $\mathbf{Z}_\text{sig}\in \mathbb{R}^{N\times D_\text{sig}}$ where the $q$-th column is $\mathbf{z}^{(q)}_\text{sig}$.

\paragraph{Output structure.}
Outputs live on a 2D grid with $D_y = H W$ locations.
We index output dimensions by $j\in \{1,\dots,D_y\}$ and associate each $j$ with a coordinate vector $\mathbf{r}_j=(u_j,v_j)^\top$ where $u =0, \ldots, H-1$ and $v =0, \ldots, W-1$.
The matrix $\mathbf{W}_\text{sig}\in \mathbb{R}^{D_y\times D_\text{sig}}$ maps latents to the output grid.
Each column $q$ is a localized squared-exponential blob with centre $(\bar{u}_q, \bar{v}_q)$ and scale $s_q$:
\begin{equation*}
  [\mathbf{W}_\text{sig}]_{j,q}= \exp\left({-}\frac{(u_j - \bar{u}_q)^2 + (v_j - \bar{v}_q)^2}{2s_q^2}\right), \quad q=1,\dots,D_\text{sig}.
\end{equation*}

\paragraph{Nuisance variation.}
For each sample $i$, draw an independent random field over the output grid:
\begin{equation*}
  \mathbf{y}_{\text{nuis},i}\sim \mathcal{N}(0,  \boldsymbol{\Sigma}_\text{nuis}),
\end{equation*}
where $\boldsymbol{\Sigma}_\text{nuis}$ is a spatial covariance across output dimensions.
We parameterize $\boldsymbol{\Sigma}_\text{nuis}$ by an RBF on grid coordinates
\begin{equation*}
  \Sigma_{\text{nuis},jj'}=\sigma^2_\text{nuis}\exp\left(-\frac{\|\mathbf{r}_j-\mathbf{r}_{j'}\|_2^2}{2 \ell_\text{nuis}^2}\right).
\end{equation*}
Stacking nuisance rows gives $\mathbf{Y}_\text{nuis}=
\begin{bmatrix} \mathbf{y}_{\text{nuis},1}^\top \\ \vdots \\ \mathbf{y}_{\text{nuis}, N}^\top
\end{bmatrix}\in \mathbb{R}^{N\times D_y}$.
In expectation, nuisance contributes only diagonal energy in example space $\mathbb{E}[\mathbf{Y}_\text{nuis} \mathbf{Y}_\text{nuis}^\top]=\mathrm{tr}(\boldsymbol{\Sigma}_\text{nuis}) \mathbf{I}_N$, which simplifies to $D_y \sigma_\text{nuis}^2$ for our RBF parameterization.

\paragraph{Inputs.}
Drawn as $\mathbf{x}_i \sim \mathcal{N}(0, \mathbf{I}_{D_x})$ and split randomly into training, validation, and test sets.

\paragraph{Settings used in main text.}
$D_y=16^2=256$, $D_x=3$, $D_\text{sig}=6$; $k(\cdot,\cdot)$ was an RBF kernel with lengthscales $\ell_q\sim\mathcal{U}(1,3)$, shared across input dimensions; $\ell_\text{nuis}=2$; $s_q\sim\mathcal{U}(1,2)$, $\bar{u}_q\sim\mathcal{U}(0,H-1)$, $\bar{v}_q\sim\mathcal{U}(0,W-1)$. For Figures~\ref{fig:synthetic-learning-curves} and \ref{fig:synthetic-latent-dim}: $(\sigma_\text{sig}, \sigma_\text{nuis}, \sigma_\epsilon)=(1,1,0.01)$. For Figure~\ref{fig:synthetic-nuisance}: $(\sigma_\text{sig}, \sigma_\epsilon)=(1,0)$ and $\sigma_\text{nuis}$ is varied.

\subsubsection{Model Settings and Hyperparameters}\label{app:synthetic:hyper}

\paragraph{Splits.}
Fixed split with sizes 1600 (train-pool), 500 (validation), 500 (test).
Training subsets in Figure~\ref{fig:synthetic-learning-curves} come from a nested $N\in\{50,100,200,400,800\}$ sweep across sampling seeds $0$--$4$.

\paragraph{Preprocessing.}
For each $(\text{seed},N)$ we fit z-score standardizers on the training subset.
We standardize per-dimension for both $\mathbf{X}$ and $\mathbf{Y}$.

\paragraph{Tuning procedure.}
Hyperparameters for both GPLFR and PCA-GP are selected by median $\mathrm{RMSE}_\text{sig}$ on a fixed 500-example validation set across the five dataset seeds of the 400-example training set and then held fixed across all shown runs.

\paragraph{Early stopping.}
All models use early stopping based on the \emph{observed} error
$\mathrm{RMSE}_\text{obs}=\sqrt{\frac{1}{N_{\text{val}}\, D_y} \sum_{i=1}^{N_{\text{val}}} \|\hat{\mathbf{y}}_i - \mathbf{y}_{i}\|_2^2}$
on the validation set, since in practice the true signal is not available for model selection.

\paragraph{GPLFR.}~

\emph{Kernel.} We use per-latent stationary ARD RBF kernels ($D_q=1\ \forall q$) with separate lengthscales and a single shared amplitude fixed to one. GPLFR was relatively insensitive to the amplitude grouping: fixing to one usually performed best, with a shared learned amplitude only marginally worse, and per-latent amplitudes again slightly worse. This is consistent with GPLFR's amplitudes being weakly identified with latent scaling.

\emph{Optimization.} We fit parameters via MAP estimation using Adam.

\emph{Priors.} We use the priors in Table~\ref{tab:synthetic-priors} (on the standardized data).

\emph{Hyperparameters.} Tuned values are listed in Table~\ref{tab:synthetic-tuned}.

\paragraph{PCA-GP.}~

\emph{Kernel.} As with GPLFR, we use per-latent ARD RBF kernels with separate lengthscales.
We learn a shared latent noise (nugget). On amplitude grouping, per-latent performed best, with a single shared or fixed amplitude significantly hurting performance.

\emph{Optimization.} We maximize the marginal likelihood using L-BFGS-B.
For the larger latent dimensions in Figure~\ref{fig:synthetic-latent-dim} we increased the number of L-BFGS-B steps from 500 to 1500 (keeping early stopping) to ensure convergence; a further increase to 3000 steps for the largest dimensions did not change the result.

\emph{Hyperparameters.} Tuned values are listed in Table~\ref{tab:synthetic-tuned}.

\begin{table}[t]
  \centering
  \begin{tabular}{@{}ll@{}}
    \toprule
    Parameter & Prior \\
    \midrule
    Kernel lengthscales & $\log \ell \sim \mathcal{N}(0, 0.3^2)$ \\
    Kernel amplitudes & $\log \eta \sim \mathcal{N}(0, 1^2)$ \\
    Observation noise & $\sigma \sim \text{HalfNormal}(0.5^2)$ \\
    \bottomrule
  \end{tabular}
  \caption{GPLFR Priors for all experiments.}
  \label{tab:synthetic-priors}
\end{table}

\begin{table}[t]
  \centering
  \begin{tabular}{@{}lll@{}}
    \toprule
    Hyperparameter & Selected & Candidates \\
    \midrule
    \multicolumn{3}{@{}l}{\textbf{GPLFR}} \\
    Lengthscale grouping & per-latent & [per-latent, shared] \\
    Amplitude grouping & fixed to 1 & [per-latent, shared, fixed to 1] \\
    Inverse-temperature ($\beta$) & 0.1 & [0.03, 0.1, 0.3] \\
    Latent noise ($\lambda$) & $10^{-5}$ & $[10^{-5}, 10^{-3}, 10^{-1}]$ \\
    Latent learning rate & 0.01 & $[0.001, 0.003, 0.01, 0.03]$ \\
    Global learning rate & 0.003 & $[0.001, 0.003, 0.01, 0.03]$ \\
    \midrule
    \multicolumn{3}{@{}l}{\textbf{PCA-GP}} \\
    Lengthscale grouping & per-latent & [per-latent, shared] \\
    Amplitude grouping & per-latent & [per-latent, shared, fixed to 1] \\
    \bottomrule
  \end{tabular}
  \caption{\emph{Synthetic benchmark:} Hyperparameters.}
  \label{tab:synthetic-tuned}
\end{table}

% \subsubsection{Training time}\label{app:synthetic:timing}

% Table~\ref{tab:synthetic-timing} reports median (5 seeds) wall-clock fitting time per run for Figure~\ref{fig:synthetic-learning-curves}.
% GPLFR fits use a fixed 5000-step Adam budget on a single NVIDIA A100 GPU, with validation checkpointing included in the reported wall time.
% PCA-GP times include PCA and CPU L-BFGS-B hyperparameter optimization (8 CPU cores, maximum of 500 iterations).

% \begin{table}[t]
%   \centering
%   \small
%   \renewcommand{\arraystretch}{1.1}
%   \caption{\emph{Synthetic benchmark:} Median (5 seeds) wall-clock fitting time per run for Figure~\ref{fig:synthetic-learning-curves}.}
%   \label{tab:synthetic-timing}
%   \begin{tabular}{@{}rcc@{}}
%     \toprule
%     $N$ & GPLFR (s) & PCA-GP (s) \\
%     \midrule
%     50  & 42 & 16 \\
%     100 & 38 & 10 \\
%     200 & 40 & 14 \\
%     400 & 48 & 20 \\
%     800 & 80 & 29 \\
%     \bottomrule
%   \end{tabular}
% \end{table}

\subsubsection{Signal versus nuisance capture}\label{app:synthetic:subspace}

Let $\mathbf{Y}_{\text{sig,test}}$ and $\mathbf{Y}_{\text{nuis,test}}$ denote the held-out signal and nuisance components of the test outputs (known from the data-generating process). For a learned output subspace with orthogonal projector $\mathbf{P}_{\hat{\mathcal{S}}}$, we define the signal and nuisance capture scores
$$C_{\text{sig}}(\hat{\mathcal{S}}) = \frac{\|\mathbf{Y}_{\text{sig,test}}\,\mathbf{P}_{\hat{\mathcal{S}}}\|_F^2}{\|\mathbf{Y}_{\text{sig,test}}\|_F^2}, \qquad C_{\text{nuis}}(\hat{\mathcal{S}}) = \frac{\|\mathbf{Y}_{\text{nuis,test}}\,\mathbf{P}_{\hat{\mathcal{S}}}\|_F^2}{\|\mathbf{Y}_{\text{nuis,test}}\|_F^2},$$
i.e., the fractions of signal and nuisance energy retained after projection onto $\hat{\mathcal{S}}$. 
Both scores rise with $D_z$ as the subspace expands, but GPLFR consistently captures more signal and less nuisance than PCA-GP.

\begin{figure}[t]
  \centering
  \includegraphics[width=0.58\linewidth]{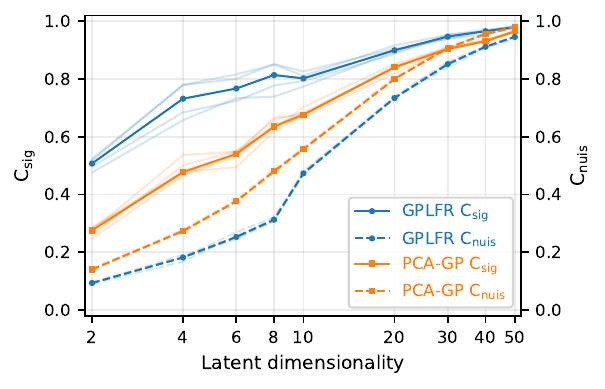}
  \caption{\emph{Synthetic benchmark:} Signal capture $C_{\text{sig}}$ (solid, left axis) and nuisance capture $C_{\text{nuis}}$ (dashed, right axis) versus latent dimensionality $D_z$, with $N=800$. GPLFR captures more signal and less nuisance than PCA-GP at every $D_z$. Bold lines show medians over five dataset seeds; faint lines show individual seeds.}
  \label{fig:synthetic-subspace-recovery}
  \vspace{-12pt}
\end{figure}

\subsubsection{GPLFR and PCA-GP for Noiseless Data}\label{app:synthetic:noiseless}

In the idealized case where the training outputs are exactly rank-$D_z$ ($\sigma^2_\text{nuis} = \sigma^2_\epsilon = 0$) and the model fits them exactly (i.e., $\mathbf{Y} = \mathbf{Z}\mathbf{W}^\top$), GPLFR and PCA necessarily span the same $D_z$-dimensional output subspace (in practice, regularization and imperfect optimization mean this holds only approximately).
However, even in this limit, PCA-GP and GPLFR are not equivalent because they treat the representation's degrees of freedom within the subspace -- i.e., any invertible transformation $(\mathbf{Z}, \mathbf{W}) \to (\mathbf{Z}\mathbf{R}, \mathbf{W}\mathbf{R}^{-\top})$ -- differently.
Since standard GP priors are not rotationally invariant, different bases within the same subspace constitute different regression tasks.

With random initialization, GPLFR may converge to a locally stable factorization that reconstructs well but yields latent coordinates less well matched to the per-latent GP priors, leading to worse generalization. PCA provides an orthogonal, variance-ordered basis that avoids such mixing, making it easier for per-latent GP hyperparameters to settle into a coherent configuration.
We can exploit PCA's good starting basis in GPLFR by initializing the latent coordinates to PCA scores. Subsequent joint optimization can then make modest, input-informed adjustments that improve kernel coherence relative to a frozen PCA basis. (With PCA initialization, we found it important to allow GPLFR to learn at least a shared GP amplitude $\eta$; otherwise, global rescaling is forced into the latents themselves, which can push the solution out of the PCA-initialized basin.)
All three effects are visible at $N = 800$, $D_z = 6$: randomly initialized GPLFR achieved an $\text{RMSE}_\text{sig}$ of $0.0117^{+0.0032}_{-0.0007}$, PCA-GP did better with $0.0091^{+0.0005}_{-0.0006}$, and PCA-initialized GPLFR slightly better still with $0.0080^{+0.0011}_{-0.0003}$ (medians $\pm\,25\text{--}75$th percentiles over 5 seeds).

We note that PCA initialization can \emph{harm} GPLFR at higher nuisance levels, presumably because it biases the latents toward a basin near the PCA solution, which generalizes poorly in those regimes.
% (GPLFR also lacks PCA-GP's ``safety mechanism'' of a learned latent noise term -- which would have identifiability issues with the learned observation noise -- perhaps making it more vulnerable.)

\subsection{Biomedical Optics: Emulating PyXOpto}\label{app:pyxopto}

\subsubsection{PyXOpto simulation details}\label{app:pyxopto:dgp}

The simulations in our subset correspond to a single-layer, semi-infinite homogeneous medium illuminated by a Gaussian beam (full-width-half-maximum $100\,\mu\mathrm{m}$).
$\mu_a$ and $\mu_s'$ are the absorption and reduced scattering coefficients, which are varied over a $21\times 21$ grid with $\mu_a\in[0,5]\,\mathrm{cm}^{-1}$ and $\mu_s'\in[5,35]\,\mathrm{cm}^{-1}$.
Scattering is modeled using a Henyey--Greenstein phase function with shape defined by the scattering anisotropy parameter $g\in \{0.1,0.3,0.5,0.7,0.9\}$ which we treat as a discrete input indexed by $s\in \{1,\dots,5\}$. 
The reflectance is collected by a radial detector $r\in[0,5]\,\mathrm{mm}$ with $D_y=500$ concentric bins.

\begin{figure}[t]
  \centering
  \includegraphics[width=0.65\linewidth]{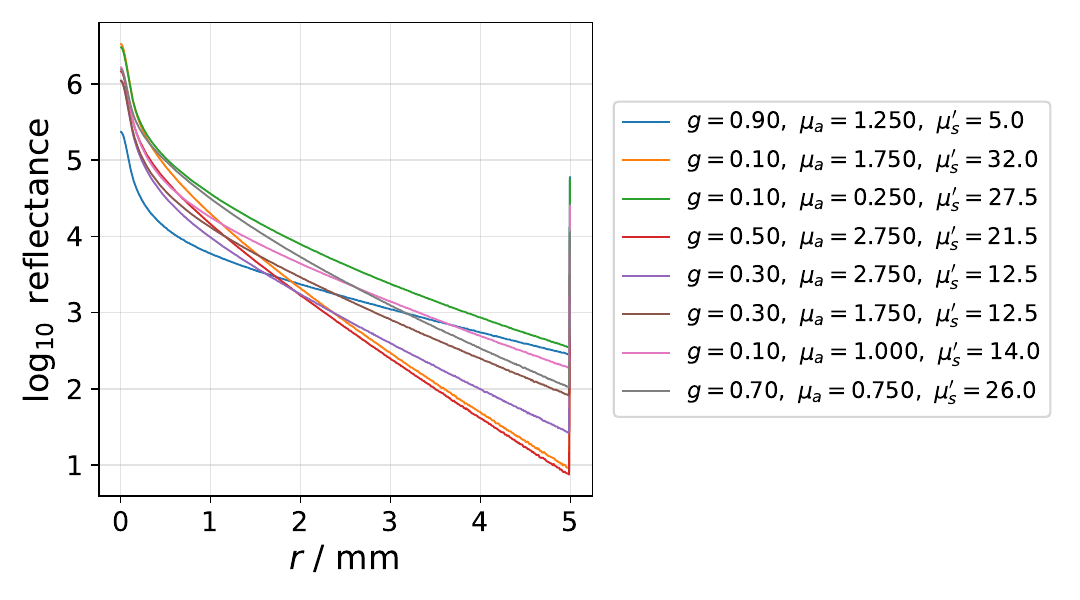}
  \caption{\emph{PyXOpto emulation:} Example reflectance curves. The spike at $5\,\mathrm{mm}$ is a binning effect.}
  \label{fig:pyxopto-example-curves}
\end{figure}

\subsubsection{Model Settings and Hyperparameters}\label{app:pyxopto:hyper}

\paragraph{Splits.}
There are 2205 examples available in total.
We use a fixed split with sizes 1205 (train-pool), 500 (validation), 500 (test) with an equal number of examples per task ($g$ value) in each partition.
For Figure~\ref{fig:pyxopto-learning-curves}, we draw balanced training subsets with $N/S$ samples per task for nested $N\in\{50,100,200,400,800\}$ across sampling seeds $0$--$4$.

\paragraph{Preprocessing.}
For each $(\text{seed},N)$ we fit standardizers on the training subset.
Inputs are z-scored per-dimension, outputs are log-transformed $y\gets \log_{10}y$ and then z-scored per-dimension.

\paragraph{Tuning procedure.}
Hyperparameters for all models are selected by median $\mathrm{RMSE}$ on a fixed 500-example validation set across five dataset seeds of the 200-example training set and then held fixed across subsequent runs.

\paragraph{Early stopping.}
All models use early stopping based on RMSE on the validation set.

\paragraph{GPLFR.}~

\emph{Kernel.} We use per-latent latent GPs ($D_q=1$) with ARD RBF kernels.
Each latent dimension has separate lengthscales and a separate amplitude variance; we found that sharing a single amplitude across latents performed marginally worse.

\emph{Coregionalization.} The $g$-coregionalization matrix $\mathbf{B}^\text{in}$ is parameterized via its Cholesky factor $\mathbf{L}$, such that $\mathbf{B}^\text{in} = \mathbf{L}\mathbf{L}^\top$.
We place an $\mathrm{LKJCholesky}(1)$ prior on $\mathbf{L}$, which ensures $\mathbf{B}^\text{in}$ is a correlation matrix.

\emph{Optimization.} We fit parameters via MAP estimation using Adam.

\emph{Hyperparameters.} Tuned values are listed in Table~\ref{tab:pyxopto-hyper}.

\paragraph{PCA-ICM.}~

\emph{Kernel.} As with GPLFR, we use per-score ARD RBF kernels with separate lengthscales and amplitudes.

\emph{Coregionalization.} We parameterize the $g$-coregionalization matrix as a correlation matrix $\mathbf{B}^\text{in} = \mathrm{corr}(\mathbf{L}\mathbf{L}^\top)$, where $\mathbf{L}$ is a lower-triangular matrix of free parameters.

\emph{Optimization.} We maximize the marginal likelihood using Adam (L-BFGS proved unstable for the correlation matrix parameters).

\emph{Priors.} We use the same priors as for the synthetic benchmark (Table~\ref{tab:synthetic-priors}).

\emph{Hyperparameters.} Tuned values are listed in Table~\ref{tab:pyxopto-hyper}.

\paragraph{PCA-MLP.}~

\emph{Architecture.} The MLP consists of two hidden layers with shared width.
The input is the concatenation $[\mathbf{x}; \mathbf{e}_s]$, where $\mathbf{e}_s$ is the one-hot encoding of task index $s$.

\emph{Training.} We minimize MSE on the PCA scores using the AdamW optimizer.

\emph{Hyperparameters.} We tuned the hidden width, learning rate, and weight decay, as detailed in Table~\ref{tab:pyxopto-hyper}.

\begin{table}[t]
\centering
\begin{tabular}{@{}lll@{}}
  \toprule
  Hyperparameter & Selected & Candidates \\
  \midrule
  \multicolumn{3}{@{}l}{\textbf{GPLFR}} \\
  Lengthscale grouping & per-latent & [per-latent, shared] \\
  Amplitude grouping & per-latent & [per-latent, shared] \\
  Inverse-temperature ($\beta$) & 0.3 & [0.03, 0.1, 0.3] \\
    Latent noise ($\lambda$) & $10^{-5}$ & $[10^{-5}, 10^{-3}, 10^{-1}]$ \\
  Latent learning rate & 0.01 & $[0.001, 0.003, 0.01, 0.03]$ \\
  Global learning rate & 0.003 & $[0.001, 0.003, 0.01, 0.03]$ \\
  \midrule
  \multicolumn{3}{@{}l}{\textbf{PCA-ICM}} \\
  Lengthscale grouping & per-latent & [per-latent, shared] \\
  Amplitude grouping & per-latent & [per-latent, shared] \\
  Learning rate & 0.01 & $[3\cdot 10^{-4}, 0.001, 0.003, 0.01, 0.03]$ \\
  \midrule
  \multicolumn{3}{@{}l}{\textbf{PCA-MLP}} \\
  Hidden widths & [256, 256] & [[32,32], [64,64], [128,128], [256,256]] \\
  Activation & SiLU & [ReLU, SiLU] \\
  Learning rate & $0.001$ & $[10^{-4}, 3\cdot 10^{-4},0.001, 0.003, 0.01]$ \\
  Weight decay & $0.001$ & $[10^{-4}, 3\cdot 10^{-4},0.001, 0.003, 0.01, 0.03]$ \\
  \bottomrule
\end{tabular}
\caption{\emph{PyXOpto emulation:} Hyperparameters.}
\label{tab:pyxopto-hyper}
\end{table}

\subsection{Emulating Exoplanet Climate Models}\label{app:exoplanet}

\subsubsection{Dataset Details}\label{app:exoplanet:data}

\paragraph{Planets.}
We focus specifically on \emph{tidally-locked aquaplanets} -- ocean-covered rocky planets in or near the habitable zone, with one hemisphere permanently facing the host star.
These are the most plentiful subclass of rocky exoplanets modelled in the literature.
We focus evaluation on planets satisfying the physical constraints in Table~\ref{tab:exoplanet-Y}.

\begin{table}[b]
  \centering
  \small
  \setlength{\tabcolsep}{5pt}
  \renewcommand{\arraystretch}{1.1}
  \begin{tabular}{lp{0.1\linewidth}}
    \toprule
    Parameter / units & Range \\
    \midrule
    Radius / Earth radii & $\left[0.7,\,1.4\right]$ \\
    Surface gravity / m\,s$^{-2}$ & $\left[6.0,\,16.0\right]$ \\
    Rotation period / days & $\left[0.1,\,1000.0\right]$ \\
    Surface pressure / bar & $\left[0.5,\,5\right]$ \\
    CO$_2$ volume fraction / \% & $\left[0,\,100\right]$ \\
    CH$_4$ volume fraction / \% & $\left[0,\,1\right]$ \\
    Incident stellar flux / W\,m$^{-2}$ & $\left[500,\,1500\right]$ \\
    Stellar temperature / K & $\left[2500,\,5800\right]$ \\
    \bottomrule
  \end{tabular}
  \caption{\emph{Exoclimate emulation:} Physical constraints defining the core domain.}
  \label{tab:exoplanet-Y}
\end{table}

\paragraph{GCMs.}
Our data spans simulations from five GCMs.
We designate ExoCAM and the UM as the \emph{target} GCMs, evaluated at test time, since they are high-fidelity models with relatively plentiful simulations.
ExoPlaSim, LFRic, and an earlier version of ExoCAM (pre-2022) serve as \emph{auxiliary} GCMs, contributing training data only.
ExoPlaSim provides the majority of the training set but is lower fidelity than the rest.
The 2022 update to ExoCAM provided an improved radiation scheme that reduced a CO$_2$-atmosphere bias in the pre-2022 setup.

\paragraph{Data sources.}
Our data is mainly sourced from existing datasets in the exoplanet science literature.
The resulting combined dataset is highly non-uniform, with clusters around community-favorite planets such as TRAPPIST-1e and Proxima Centauri b, and each constituent study tending to vary only a few parameters.
To mitigate this, we ran $357$ bespoke simulations chosen to fill gaps in the input space using a weighted coverage design.
A summary of the dataset is given in Table~\ref{tab:exoplanet-W}.

\paragraph{Train--test split.}
Train--test splitting is based solely on the focus set examples (target GCM $\cap$ core domain).
All remaining examples (auxiliary GCMs or outside the core domain) are included in the training set to improve predictions via cross-model transfer and broader input-space coverage.
Out-of-domain simulations typically violate the core domain constraints in only one or two input dimensions, so can remain informative for anchoring the response surface near the domain boundaries.
To prevent leakage, duplicated planets (same continuous parameters, different GCM) are never split across train and test.

\begin{table*}[t]
  \centering
  \footnotesize
  \setlength{\tabcolsep}{5pt}
  \renewcommand{\arraystretch}{1.1}
  \begin{tabular}{@{}lccp{0.60\textwidth}@{}}
    \toprule
    GCMs & \shortstack{Within\\core domain} & Additional & Data sources \\
    \midrule
    \multicolumn{4}{@{}l}{\textbf{Target GCMs}} \\
    UM & \textbf{228} & 31 & \cite{mak3DSimulations2024, sergeevTRAPPIST1Habitable2022, stevensonThousandWorlds2026} \\
    ExoCAM & \textbf{87} & 7 & \cite{hammondClimatesThermal2025, haqq-misraSparseAtmospheric2022, sergeevTRAPPIST1Habitable2022,wolfChemistryClimate2025,woodwardExoCAMInPrep,stevensonThousandWorlds2026}\\
    \midrule
    \multicolumn{4}{@{}l}{\textbf{Auxiliary GCMs}} \\
    ExoCAM pre-2022 & 113 & 47 & \cite{komacekAtmosphericCirculation2019, kopparapuINNEREDGE2016, kopparapuHabitableMoist2017, wolfSimulatedPhasedependent2019, wolfAssessingHabitability2017, suissaFirstHabitablezone2020} \\
    LFRic & 14 & 5 & \cite{haqq-misraSparseAtmospheric2022,stevensonThousandWorlds2026}\\
    ExoPlaSim & 426 & 776 & \cite{macdonaldClimateTransition2025, paradiseClimateDiversity2021, paradiseExoPlaSimExtending2022, paradiseFundamentalChallenges2022,stevensonThousandWorlds2026} \\
    \bottomrule
  \end{tabular}
\caption{\emph{Exoclimate emulation:} GCM dataset summary. \emph{Target} GCMs are evaluated at test time; \emph{auxiliary} GCMs contribute training data only. \emph{Within core domain} counts simulations satisfying the physical constraints in Table~\ref{tab:exoplanet-Y}; \emph{additional} counts simulations outside these constraints that are included in training only. Train--test splitting is based solely on the 315 \emph{focus set} examples (target GCM $\cap$ core domain); the remaining 1419 examples are included to improve predictions via cross-model and cross-domain transfer.}
\label{tab:exoplanet-W}
\end{table*}

\subsubsection{Preprocessing}\label{app:exoplanet:preprocessing}

\paragraph{Grid.}
Each GCM produces output on its own native grid.
We regrid all fields onto a common $32\times64$ Gaussian latitude--longitude grid (with 10 pressure levels for 3D variables): horizontal interpolation is bilinear, vertical interpolation is in log-pressure, and any field with partially missing values is treated as fully unobserved.
This grid's nodes support an exact spherical harmonic transform up to total wavenumber 21 (a ``T21'' grid).

The 10 pressure levels are defined as relative isobars $\sigma_k = (P_k - P_\text{top}) / (f_\text{bottom} P_0 - P_\text{top})$, where $P_0$ is the input surface pressure, $P_\text{top} = 10\,\text{mbar}$, and $f_\text{bottom} = 0.95$ lifts the lowest level above near-surface pressure fluctuations.
The $\sigma_k$ are spaced between 0 and 1 according to a fourth-order polynomial that slightly increases resolution near the top and bottom of the atmosphere.

\paragraph{Inputs.}
Rotation period and surface pressure are log-transformed.
Gas volume fractions are transformed via $x\gets\operatorname{asinh}(x/s)$, where $s$ is a fixed pivot per species (CO$_2$: $10^{-6}$; CH$_4$: $10^{-8}$), chosen so that the transformation is approximately logarithmic at climatically significant concentrations and linear near zero.
All inputs are then z-scored over the training set.

\paragraph{Outputs.}
Specific humidity is log-transformed and cloud fraction is smoothed-logit-transformed: $c\mapsto\operatorname{logit}((c+\varepsilon)/(1+2\varepsilon))$ with $\varepsilon=2\times10^{-3}$ (selected via a small sweep); predictions are clamped back to $[0,1]$ after inversion.
Each horizontal field is then expanded in the T21 spherical harmonic basis.
The resulting coefficient vectors are centered and scaled per-field: for field $k$, we compute the training-set mean $\bar{\mathbf{a}}^{(k)}$ and the root-mean anomaly energy
\begin{equation*}
  \sigma^{(k)}=\sqrt{\frac{1}{N}\sum_{i=1}^N\|\mathbf{a}_i^{(k)}-\bar{\mathbf{a}}^{(k)}\|_2^2}
\end{equation*}
and set $\mathbf{y}_i^{(k)}=(\mathbf{a}_i^{(k)}-\bar{\mathbf{a}}^{(k)})/\sigma^{(k)}$.
Dividing by $\sigma^{(k)}$ equalizes the total anomaly variance across fields.
Predictions are denormalized by inverting these steps before evaluation.

\paragraph{Equatorial symmetry.}
All planets in our dataset have symmetric forcing about the equator, so their time-mean climates should be equatorially symmetric in all fields except meridional wind $v$, which is antisymmetric.
In practice, some simulations exhibit residual asymmetry due to finite time-averaging windows and, in some cases, spontaneous symmetry breaking or amplified numerical asymmetries.
We treat these as artifacts (or at least beyond the emulator's modeling scope) and enforce symmetry by zeroing the complementary spherical harmonic coefficients: since $Y_l^m(\pi-\theta,\phi)=(-1)^{l+m}Y_l^m(\theta,\phi)$, symmetry corresponds to retaining only $l+m$ even coefficients and antisymmetry to retaining only $l+m$ odd.
In spatial terms, this is equivalent to averaging the two hemispheres (for symmetric fields) or taking half their difference (for $v$).

\subsubsection{GPLFR Configuration}\label{app:exoplanet:gplfr}

\paragraph{Output-space mean function.}
A stationary, zero-mean GP encoder can capture global trends within the training domain, but out-of-domain predictions revert toward the prior mean.
Following standard emulator practice, we separate a low-capacity parametric trend from the GPLFR residual model by fitting and subtracting a linear mean function in output space before the GP is trained.
Concretely, we regress each output dimension on the standardized inputs plus a one-hot encoding of GCM identity via ridge regression with a small penalty ($\lambda_\text{ridge}=10^{-3}$) for conditioning, and replace the outputs with the residuals.
At prediction time, the linear trend is added back to posterior predictive samples.

\paragraph{Input kernel.}
We use an ICM kernel $k((\mathbf{x},s),(\mathbf{x}',s'))=k_x(\mathbf{x},\mathbf{x}')\,B^\text{in}_{ss'}$.
The continuous component $k_x$ is an ARD Mat\'ern-$\tfrac{5}{2}$ kernel on the standardized planet properties (the kernel family was selected from RBF, Mat\'ern-$\tfrac{3}{2}$, and Mat\'ern-$\tfrac{5}{2}$ independently per model class during cross-validation; see Appendix~\ref{app:exoplanet:training}).
The coregionalization matrix is decomposed as $\mathbf{B}^\text{in}=\mathrm{diag}(\mathbf{r})\,\mathbf{R}\,\mathrm{diag}(\mathbf{r})$, where $\mathbf{R}\in\mathbb{R}^{S\times S}$ is a correlation matrix with an $\mathrm{LKJCholesky}(1)$ prior and $\mathbf{r}\in\mathbb{R}^S_{>0}$ are per-GCM latent amplitude scales, constrained to have unit geometric mean to remove a global scale redundancy with $k_x$.
This separates two interpretable quantities: $\mathbf{R}$ captures how similarly two GCMs respond to changes in planet properties, and $r_s$ captures the overall sensitivity of GCM $s$ relative to the others.

\paragraph{Missing data.}
Different GCMs use different vertical grids, not all simulations extend to the same altitude, and some studies do not output all variables we emulate.
Additionally, GCMs that use height-based rather than pressure-based vertical coordinates can have pressure fluctuations that leave our lowest or highest isobars only partially observed across the spatial grid; to avoid introducing bias, we treat such partially missing fields as fully unobserved.
After interpolation onto the canonical 10-level pressure grid, this leaves a structured pattern of missingness: some examples lack data at upper or lower pressure levels, and others lack entire fields.
We handle this within the collapsed decoder likelihood by restricting each output dimension's contribution to the examples where it is observed, assuming missingness is independent of the unobserved values given the inputs.
Output dimensions sharing the same set of observed examples are grouped so that the Woodbury inversion is computed once per missingness pattern, adding a negligible $O(PD_z^3)$ cost where $P$ is the number of distinct patterns.

\paragraph{Model fitting.}
All parameters are fit jointly via MAP estimation using Adam. 

\subsubsection{PPCA-ICM}\label{app:exoplanet:ppca}

PPCA-ICM is a compress-then-predict baseline analogous to PCA-GP, but one that (i) replaces PCA with probabilistic PCA (PPCA), and (ii) replaces independent per-score GPs with a multi-task GP using an intrinsic coregionalization model (ICM) over GCM labels.
It operates in the same normalized spectral-coefficient space and uses the same equatorial symmetry masking, missing-field conventions, and output-space mean function as GPLFR.

\paragraph{Stage 1: PPCA compression.}
Each example's observed coefficient vectors are modelled as a linear function of a shared latent $\mathbf{z}_i\in\mathbb{R}^{D_z}$ with field-specific loadings and means and a shared isotropic noise variance $\sigma^2$.
Missing fields are handled by omitting their likelihood terms.
Parameters are fit by maximum likelihood via 50 iterations of expectation-maximization, which also yields posterior score estimates $\mathbf{z}_i$ for each training example.

\paragraph{Stage 2: ICM-GP regression.}
The PPCA scores are regressed against inputs $(\mathbf{x},s)$ using independent GPs with an ICM kernel: $k((\mathbf{x},s),(\mathbf{x}',s'))=k_x(\mathbf{x},\mathbf{x}')\,B^\text{in}_{ss'}$, where $k_x$ is an ARD Mat\'ern-$\tfrac{5}{2}$ kernel on the continuous planet properties (selected independently from the same kernel-family candidates as GPLFR) and $\mathbf{B}^\text{in}\in\mathbb{R}^{S\times S}$ is a coregionalization matrix across GCMs, using the same correlation-plus-scales parameterization as GPLFR.
We compared a shared-kernel regime (all scores drawn from a GP with common lengthscales, amplitudes, and $\mathbf{B}^\text{in}$) against a per-component regime (separate lengthscales and amplitudes per score, with shared $\mathbf{B}^\text{in}$); the shared-kernel regime performed better and is used throughout.
Kernel hyperparameters are fit by maximizing the GP marginal likelihood using Adam with learning rate $10^{-3}$.

\paragraph{Prediction.}
Ensemble members are generated by sampling latent scores from the GP predictive distribution, decoding each through the PPCA loadings and means, and adding PPCA noise.
The resulting spectral coefficients are mapped back to spatial fields via the same inverse preprocessing as GPLFR.

\subsubsection{SV-LMC}\label{app:exoplanet:svlmc}

SV-LMC is a sparse variational LMC (Section~\ref{sec:lmc:def}) with 1D latent groups: $D_z$ independent latent GPs, each with its own kernel, are mixed into the $D_y$ outputs by a learned unrestricted loading matrix $\mathbf{W}\in\mathbb{R}^{D_y\times D_z}$, so each latent contributes a rank-1 coregionalization matrix $\mathbf{B}_q=\mathbf{w}^{(q)}\mathbf{w}^{(q)\top}$ \citep{tehSemiparametricLatent2005,wilkFrameworkInterdomain2020}.
It operates in the same normalized spectral-coefficient space as GPLFR and PPCA-ICM, with the same equatorial symmetry masking, missing-field conventions, and output-space mean function.
Each latent kernel is an ARD Mat\'ern-$\tfrac{5}{2}$ over the continuous planet properties multiplied by a coregionalization factor over the GCM label shared across latents, matching the input-side treatment of the other GP models; the model differs from them in placing its learned covariance structure on the output side, through per-latent kernels and free loadings.

\paragraph{Inference.}
We use minibatch variational inference \citep{hensmanGaussianProcesses2013,nguyenCollaborativeMultioutput2014,wilkFrameworkInterdomain2020} with $M=256$ inducing points per latent (fixed at a random subset of training inputs) and independent full-covariance Gaussian variational posteriors per latent process, implemented in GPyTorch \citep{gardnerGPyTorch2018}.
The ELBO is optimized with Adam (batch size 256, learning rate $3\times10^{-3}$) in double precision. The loading matrix is initialized from PPCA loadings.
Missing fields contribute no likelihood terms (which the ELBO handles exactly).

\paragraph{Prediction.}
Ensemble members are generated by sampling the latent marginals at test inputs, mixing through $\mathbf{W}$, and adding likelihood noise; spectral coefficients are mapped back to spatial fields via the same inverse preprocessing as GPLFR.

\subsubsection{Deterministic Baselines}\label{app:exoplanet:baselines}

\paragraph{PPCA-MLP.}
PPCA-MLP shares PPCA-ICM's compression stage (Section~\ref{app:exoplanet:ppca}, Stage~1), using the same PPCA scores as regression targets.
The MLP consists of two hidden layers with shared width, taking as input the concatenation $[\mathbf{x};\, \mathbf{e}_s]$ where $\mathbf{e}_s$ is the one-hot encoding of the GCM label.
We minimize MSE on the PPCA scores using AdamW.
Predictions are decoded through the PPCA loadings and means without noise sampling.

\paragraph{kNN.}
For each test input, we find the $k$ nearest training examples in standardized continuous input space using Euclidean distance.
GCM identity is encoded as a scaled one-hot vector appended to the input, so that different GCMs contribute an additional distance penalty $\lambda\sqrt{2}$, encouraging same-GCM neighbours while still permitting cross-GCM matches.
The prediction is the uniform average of the neighbours' spatial output fields, computed after variable-specific transforms (log for humidity, smoothed logit for cloud fraction, identity for other variables) and mapped back to physical units; missing fields are averaged only over neighbours where they are observed.
For fair comparison, the same spectral truncation and equatorial symmetry enforcement used by GPLFR are applied to predictions and test targets at evaluation.

\paragraph{Training mean.}
The prediction for every test input is the global per-field mean over the entire training set, computed after the same variable-specific transforms and mapped back to physical units.
We use a global rather than per-GCM mean because GCMs cover different regions of input space, so per-GCM means would conflate simulator differences with input-space sampling bias.

\subsubsection{Energy score estimator}\label{app:exoplanet:energy-score-estimator}

Given posterior predictive samples $\{\mathbf{y}^{[m]}\}_{m=1}^M$, we estimate the energy score as
\begin{equation*}
  \begin{aligned}
  \widehat{\mathrm{ES}}\big(\mathbf{y};\{\mathbf{y}^{[m]}\}_{m=1}^M\big)
  &=\frac{1}{M}\sum_{m=1}^M \|\mathbf{y}^{[m]}-\mathbf{y}\|_G -\frac{1}{2M(M-1)}
  \sum_{\substack{m,m'=1\\m\ne m'}}^M
  \|\mathbf{y}^{[m]}-\mathbf{y}^{[m']}\|_G .
  \end{aligned}
\end{equation*}
The results shown use $M=64$.

\subsubsection{Training and model selection}\label{app:exoplanet:training}

\begin{table}[t]
  \centering
  \setlength{\tabcolsep}{6pt}
  \renewcommand{\arraystretch}{1.1}
  \begin{tabular}{@{}ll@{}}
    \toprule
    Hyperparameter & Selected \\
    \midrule
	    \multicolumn{2}{@{}l}{\textbf{GPLFR}} \\
	    Latent dimension ($D_z$) & 150 \\
	    Input kernel family & Mat\'ern-$\tfrac{5}{2}$ \\
	    Lengthscale grouping & shared \\
	    Amplitude grouping & shared \\
	    Inverse-temperature ($\beta$) & 0.1 \\
	    Latent noise ($\lambda$) & 0.1 \\
	    Latent learning rate & 0.1 \\
	    Global learning rate & 0.3 \\
    \midrule
    \multicolumn{2}{@{}l}{\textbf{PPCA-ICM}} \\
    Latent dimension ($D_z$) & 150 \\
    Input kernel family & Mat\'ern-$\tfrac{5}{2}$ \\
    Lengthscale grouping & shared \\
    Amplitude grouping & shared \\
    Learning rate & $10^{-3}$ \\
    \midrule
    \multicolumn{2}{@{}l}{\textbf{SV-LMC}} \\
    Latent dimension ($D_z$) & 150 \\
    Input kernel family & Mat\'ern-$\tfrac{5}{2}$ \\
    Inducing points ($M$) & 256 \\
    Learning rate & $3\times10^{-3}$ \\
    Batch size & 256 \\
    \midrule
    \multicolumn{2}{@{}l}{\textbf{PPCA-MLP}} \\
    Latent dimension ($D_z$) & 50 \\
    Hidden width & 512 \\
    Learning rate & $3\times10^{-4}$ \\
    Weight decay & $10^{-5}$ \\
    \midrule
    \multicolumn{2}{@{}l}{\textbf{kNN}} \\
    Neighbours ($k$) & 3 \\
    GCM distance penalty ($\lambda$) & 3 \\
    \bottomrule
  \end{tabular}
  \caption{\emph{Exoclimate emulation:} Hyperparameters.}
  \label{tab:exoplanet-X}
\end{table}

\paragraph{Cross-validation and hyperparameter selection.}
Hyperparameters for GPLFR, PPCA-ICM, PPCA-MLP, and SV-LMC are selected via 3-fold cross-validation.
Only focus set examples are assigned to validation folds; all other training examples -- those from auxiliary GCMs and/or outside the core domain -- are included in every fold's training partition.
To prevent leakage, duplicated planets (same continuous parameters, different GCM) are grouped and assigned to folds as a unit.
For each candidate setting, we train on each fold and evaluate RMSE in normalized spectral-coefficient space on its held-out set.
We select the setting with the best mean performance across folds, then refit once on the full training set. Both GPLFR and PPCA-ICM performed well out to the highest latent dimensionality tested ($D_z=200$), with the chosen $D_z=150$ lying in the performance plateau of each. PPCA-MLP peaked around $D_z=50$ and tended to overfit at higher $D_z$.
For SV-LMC we tuned the number of latent processes $D_z\in\{10,50,150\}$; $D_z=150$ performed best.
The kNN hyperparameters were selected via 5-fold CV on the training set.
The selected hyperparameters are shown in Table~\ref{tab:exoplanet-X}.

\paragraph{Early stopping.}
GPLFR, PPCA-ICM, PPCA-MLP, and SV-LMC use early stopping based on validation RMSE in normalized spectral-coefficient space, averaged with equal weight per variable group, during cross-validation.
When refitting on the full training set for the final model, we train for the median best-validation step across folds.

\subsubsection{Training time}\label{app:exoplanet:timing}

Table~\ref{tab:exoplanet-timing} reports approximate wall-clock fitting times for the models in the exoclimate experiment in Section~\ref{sec:exoplanet} (not counting hyperparameter tuning time). All runs used a single NVIDIA H100 PCIe GPU.

\begin{table}[t]
  \centering
  \small
  \renewcommand{\arraystretch}{1.1}
  \caption{\emph{Exoclimate emulation:} Approximate wall-clock fitting times.}
  \label{tab:exoplanet-timing}
  \begin{tabular}{@{}lc@{}}
    \toprule
    Method & Fitting time (minutes) \\
    \midrule
    PPCA-MLP & 0.2 \\
    PPCA-ICM & 0.5 \\
    GPLFR    & 0.6 \\
    SV-LMC  & 4.4 \\
    \bottomrule
  \end{tabular}
\end{table}

\subsubsection{Bootstrap intervals for RMSE results}\label{app:exoplanet:uncertainty-scores}

Table~\ref{tab:exoplanet-bootstrap} reports 95\% paired bootstrap intervals on the RMSE results: 1000 resamples of the 80 test planets, each drawn with replacement and applied jointly to every method and training seed, with per-method scores averaged over training seeds within each resample.
On the per-variable differences from GPLFR (Table~\ref{tab:exoplanet-bootstrap-delta}), 21 of 24 method--variable intervals exclude zero. The exceptions are PPCA-ICM on cloud fraction and specific humidity and SV-LMC on north--south wind, although GPLFR is still clearly better on average in all three cases.
The training-seed standard deviations are small compared to these intervals (median ratio of seed standard deviation to interval width $\approx 0.04$), indicating that test-set composition is the dominant source of uncertainty in the reported scores.

\begin{table*}[htbp]
  \centering
  \scriptsize
  \setlength{\tabcolsep}{8pt}
  \renewcommand{\arraystretch}{1.05}
  \begin{tabular}{@{}p{0.3\textwidth}rrrr@{}}
    \toprule
    Variable & GPLFR & PPCA-ICM & SV-LMC & PPCA-MLP \\
    \midrule
    Surface temperature / K & 9.32\,[7.52, 11.5] & 10.4\,[8.39, 12.8] & 10.3\,[8.57, 12.4] & 11.5\,[9.66, 13.7] \\
    Temperature (3D) / K & 7.76\,[6.47, 9.50] & 8.73\,[7.24, 10.6] & 9.66\,[8.44, 11.2] & 10.2\,[8.62, 12.0] \\
    Specific humidity (3D) / dex & 0.379\,[0.319, 0.454] & 0.402\,[0.340, 0.477] & 0.469\,[0.404, 0.541] & 0.518\,[0.455, 0.591] \\
    Cloud fraction (3D) / 1 & 0.0504\,[0.0417, 0.0594] & 0.0529\,[0.0433, 0.0625] & 0.0569\,[0.0475, 0.0660] & 0.0591\,[0.0498, 0.0682] \\
    East--west wind (3D) / m\,s$^{-1}$ & 8.51\,[7.38, 9.71] & 9.63\,[8.41, 10.9] & 9.67\,[8.41, 10.9] & 10.4\,[9.21, 11.7] \\
    North--south wind (3D) / m\,s$^{-1}$ & 3.70\,[3.29, 4.22] & 4.31\,[3.77, 4.90] & 3.88\,[3.50, 4.32] & 4.47\,[4.05, 4.97] \\
    Absorbed shortwave radiation / W\,m$^{-2}$ & 19.6\,[17.4, 21.8] & 23.5\,[20.1, 27.4] & 22.5\,[20.6, 24.5] & 28.7\,[25.5, 32.1] \\
    Outgoing longwave radiation / W\,m$^{-2}$ & 12.8\,[11.2, 14.6] & 15.3\,[13.5, 17.3] & 16.2\,[14.8, 17.6] & 16.5\,[14.8, 18.2] \\
    \bottomrule
  \end{tabular}
  \caption{\emph{Exoclimate emulation:} RMSE with 95\% paired bootstrap intervals (1000 resamples of the 80 test planets, drawn with replacement and applied jointly to every method and training seed; per-method scores are averaged over training seeds within each resample).}
  \label{tab:exoplanet-bootstrap}
\end{table*}

\begin{table*}[htbp]
  \centering
  \scriptsize
  \setlength{\tabcolsep}{8pt}
  \renewcommand{\arraystretch}{1.05}
\begin{tabular}{@{}p{0.3\textwidth}rrr@{}}
    \toprule
    Variable & PPCA-ICM & SV-LMC & PPCA-MLP \\
    \midrule
    Surface temperature / K & 1.05\,[0.190, 1.92] & 0.941\,[0.121, 1.86] & 2.22\,[1.12, 3.39] \\
    Temperature (3D) / K & 0.977\,[0.301, 1.72] & 1.88\,[1.10, 2.69] & 2.45\,[1.49, 3.50] \\
    Specific humidity (3D) / dex & 0.0229\,[-0.0107, 0.0549] & 0.0885\,[0.0492, 0.133] & 0.138\,[0.0996, 0.178] \\
    Cloud fraction (3D) / 1 & 0.00255\,[-0.000686, 0.00575] & 0.00654\,[0.00344, 0.00967] & 0.00876\,[0.00636, 0.0111] \\
    East--west wind (3D) / m\,s$^{-1}$ & 1.12\,[0.484, 1.77] & 1.12\,[0.483, 1.89] & 1.90\,[1.48, 2.32] \\
    North--south wind (3D) / m\,s$^{-1}$ & 0.611\,[0.328, 0.907] & 0.174\,[-0.0473, 0.383] & 0.771\,[0.547, 0.999] \\
    Absorbed shortwave radiation / W\,m$^{-2}$ & 4.02\,[2.04, 6.48] & 2.95\,[1.30, 4.54] & 9.13\,[6.65, 11.8] \\
    Outgoing longwave radiation / W\,m$^{-2}$ & 2.47\,[1.49, 3.48] & 3.30\,[2.16, 4.32] & 3.65\,[2.44, 4.83] \\
    \bottomrule
  \end{tabular}
  \caption{\emph{Exoclimate emulation:} RMSE differences from GPLFR with 95\% paired bootstrap intervals. Positive means worse than GPLFR.}
  \label{tab:exoplanet-bootstrap-delta}
\end{table*}

\begin{figure*}[t]
  \centering
  \begin{minipage}{0.99\textwidth}
    \large\textbf{\ (a)}\par
    \includegraphics[width=\textwidth]{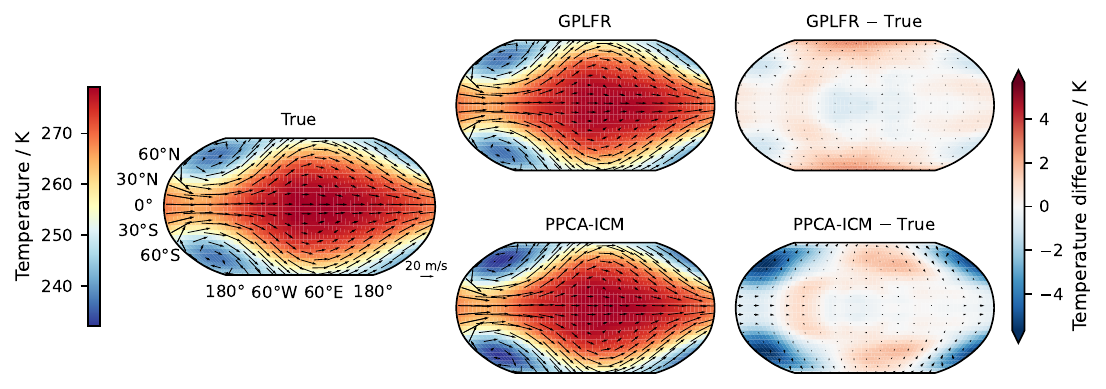}
  \end{minipage}
  \vspace{5pt}

  \begin{minipage}{1.0\textwidth}
    \large\textbf{\ (b)}\par
    \includegraphics[width=\textwidth]{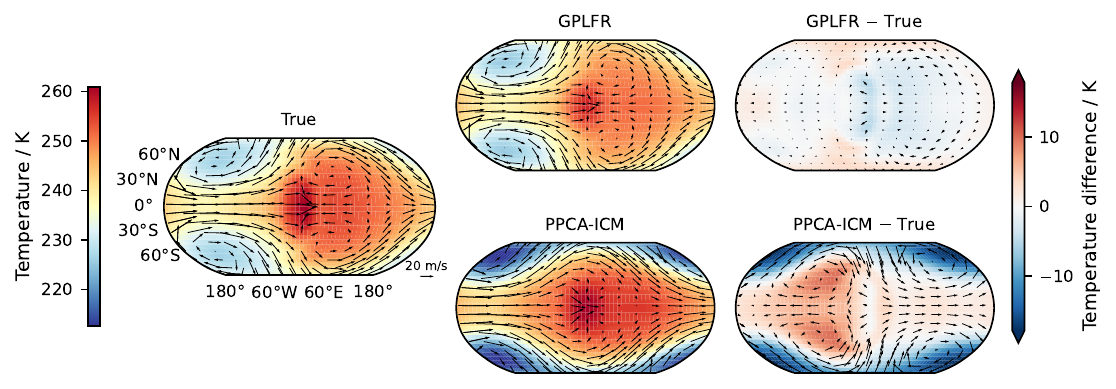}
  \end{minipage}\hspace{-3pt}
  \caption{\emph{Exoclimate emulation:} Air temperature maps with
  superimposed wind vectors at relative isobar $\sigma_3\approx 0.72$
  (this would be around the mid-troposphere on Earth) for two test planets. In \textbf{(a)},
  both models capture the broad spatial structure; GPLFR is
  near-identical to the GCM output, while PPCA-ICM overestimates the
  day--night temperature contrast. In \textbf{(b)}, a harder case,
  GPLFR still recovers the temperature and wind structure, including the
  cold vortices in the western hemisphere, while PPCA-ICM severely
  overestimates how cold they are and fails to reproduce the vortex
  circulation.}
  \label{fig:exoplanet-example}
\end{figure*}

\begin{figure*}[t]
  \centering
  \begin{minipage}{1.0\textwidth}
    \large\textbf{\ (a)}\par
    \includegraphics[width=\textwidth]{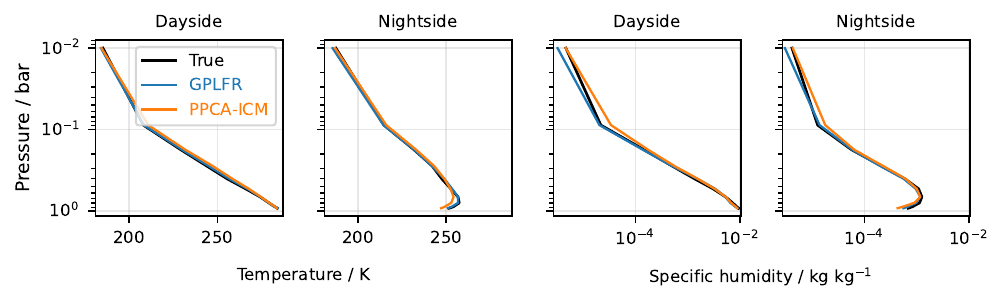}
  \end{minipage}
  \vspace{5pt}

  \begin{minipage}{1.0\textwidth}
    \large\textbf{\ (b)}\par
    \includegraphics[width=\textwidth]{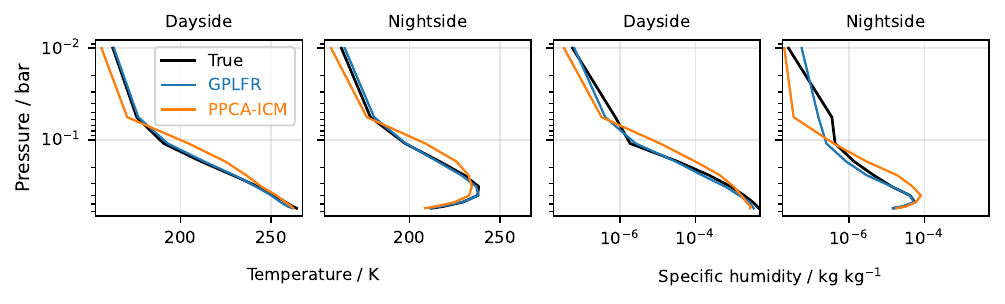}
  \end{minipage}
  \caption{\emph{Exoclimate emulation:} Dayside and nightside vertical
  profiles of area-weighted mean air temperature and specific humidity for the same two test planets as Figure~\ref{fig:exoplanet-example}. 
  In \textbf{(a)}, both models track the GCM profiles well, with noticeable deviations only at the top two levels. 
  In \textbf{(b)}, the harder case, GPLFR closely tracks the temperature profiles, while PPCA-ICM deviates significantly. Both models struggle more with specific humidity on this planet, but GPLFR is still notably closer than PPCA-ICM.
  }
  \label{fig:exoplanet-profiles}
\end{figure*}

\subsubsection{Example Predictions}\label{app:exoplanet:examples}
Figures~\ref{fig:exoplanet-example} and~\ref{fig:exoplanet-profiles}
compare GPLFR and PPCA-ICM against GCM ground truth for a small
subset of the full model output on two test planets: one where both
models perform well and one where PPCA-ICM degrades noticeably.
Figure~\ref{fig:exoplanet-example} shows spatial maps of temperature
and wind, and Figure~\ref{fig:exoplanet-profiles} shows vertical
profiles of temperature and humidity.

\clearpage
\subsubsection{Additional Results}\label{app:exoplanet:additional}

Here we expand the main-text results (Tables~\ref{tab:exoplanet-energy-summary} and~\ref{tab:exoplanet-rmse-summary}) by reporting per-level breakdowns (Tables~\ref{tab:exoplanet-Z} and~\ref{tab:exoplanet-AA}) and two additional metrics (Tables~\ref{tab:exoplanet-AB} and~\ref{tab:exoplanet-AC}).

\paragraph{Anomaly correlation coefficient (ACC).}
ACC measures agreement in spatial structure between prediction $\hat{\mathbf{y}}$ and truth $\mathbf{y}$, ignoring differences in global mean and amplitude.
For a single field,
\begin{equation*}
  \text{ACC}(\hat{\mathbf{y}},\mathbf{y})=\frac{\langle\hat{\mathbf{y}}^\circ,\mathbf{y}^\circ\rangle_G}{\|\hat{\mathbf{y}}^\circ\|_G\,\|\mathbf{y}^\circ\|_G},
  \quad
  \mathbf{u}^\circ\equiv\mathbf{u}-\bar{u}\mathbf{1},
  \quad
  \bar{u}=\langle\mathbf{1},\mathbf{u}\rangle_G/\langle\mathbf{1},\mathbf{1}\rangle_G,
\end{equation*}
where $\langle \mathbf{u}, \mathbf{v}\rangle_G \equiv \mathbf{u}^\top\mathbf{G}\mathbf{v}$ is the area-weighted inner product.
ACC $=1$ indicates perfect spatial correlation; for a random prediction, $\mathbb{E}[\text{ACC}]=0$.

GPLFR achieves the highest mean ACC on all variables except absorbed shortwave radiation, where it and SV-LMC are tied at the reported precision (both near one) (Table~\ref{tab:exoplanet-AB}). 
Its weakest results are for cloud fraction and the uppermost pressure levels: kNN beats the learned models at the top two levels of specific humidity and north--south wind, at the top level of east--west wind, and at some of the cloud fraction levels. This is likely because cloud fraction and the uppermost pressure levels of many of these variables have relatively high power at high spatial frequencies, some of which is removed by the spectral truncation used by these models.
All models achieve near-perfect ACC for absorbed shortwave radiation, reflecting the strong dependence of this field on substellar point geometry (which is consistent across examples).

\paragraph{Spread--skill ratio (SSR).}
SSR is a first-order diagnostic of ensemble calibration.
We define it as
\begin{equation*}
  \text{SSR}=\sqrt{\frac{\sum_i\text{Spread}_i^2}{\sum_i\text{MSE}_i}},
\end{equation*}
where for each test example $i$,
\begin{equation*}
  \text{Spread}_i^2=\frac{1}{M-1}\sum_{m=1}^M\|\mathbf{y}^{[m]}_i-\hat{\mathbf{y}}_i\|_G^2,\qquad\text{MSE}_i=\|\hat{\mathbf{y}}_i-\mathbf{y}_i\|_G^2,
\end{equation*}
and $\hat{\mathbf{y}}_i$ is the ensemble mean.
A well-calibrated ensemble has SSR $\approx 1$; values below 1 indicate overconfidence (too little spread) and values above 1 indicate underconfidence.

GPLFR is closer to SSR $=1$ than PPCA-ICM for 43 out of 53 fields, and has the smallest mean deviation from SSR $=1$ of the three probabilistic models (0.32, against 0.38 for SV-LMC and 0.44 for PPCA-ICM; Table~\ref{tab:exoplanet-AC}).
GPLFR and PPCA-ICM tend toward overconfidence (SSR $<1$ on 51 of 53 fields each). 
SV-LMC varies between overconfidence and underconfidence across different variables, with SSR $> 1$ on 28 of 53 fields; it is the best calibrated on winds.

\begin{table*}[t]
  \centering
  \scriptsize
  \setlength{\tabcolsep}{5pt}
  \renewcommand{\arraystretch}{1.1}
  \begin{tabular}{@{}p{0.3\textwidth}rrr@{}}
    \toprule
    Variable & GPLFR & PPCA-ICM & SV-LMC \\
    \midrule
    Surface temperature / K & \textbf{7.17} & 8.25 & 9.28 \\
    Temperature (3D) / K & \textbf{6.01} & 7.03 & 8.52 \\
    \textcolor{gray}{Temperature 0} & \textcolor{gray}{\textbf{6.30}} & \textcolor{gray}{7.13} & \textcolor{gray}{8.08} \\
    \textcolor{gray}{Temperature 1} & \textcolor{gray}{\textbf{6.22}} & \textcolor{gray}{7.07} & \textcolor{gray}{7.95} \\
    \textcolor{gray}{Temperature 2} & \textcolor{gray}{\textbf{5.82}} & \textcolor{gray}{6.74} & \textcolor{gray}{7.63} \\
    \textcolor{gray}{Temperature 3} & \textcolor{gray}{\textbf{5.34}} & \textcolor{gray}{6.24} & \textcolor{gray}{7.33} \\
    \textcolor{gray}{Temperature 4} & \textcolor{gray}{\textbf{4.92}} & \textcolor{gray}{5.72} & \textcolor{gray}{7.30} \\
    \textcolor{gray}{Temperature 5} & \textcolor{gray}{\textbf{4.73}} & \textcolor{gray}{5.75} & \textcolor{gray}{7.52} \\
    \textcolor{gray}{Temperature 6} & \textcolor{gray}{\textbf{4.83}} & \textcolor{gray}{6.08} & \textcolor{gray}{8.26} \\
    \textcolor{gray}{Temperature 7} & \textcolor{gray}{\textbf{6.95}} & \textcolor{gray}{8.05} & \textcolor{gray}{10.1} \\
    \textcolor{gray}{Temperature 8} & \textcolor{gray}{\textbf{8.44}} & \textcolor{gray}{10.1} & \textcolor{gray}{11.3} \\
    \textcolor{gray}{Temperature 9} & \textcolor{gray}{\textbf{6.54}} & \textcolor{gray}{7.43} & \textcolor{gray}{9.73} \\
    Specific humidity (3D) / dex & \textbf{0.660} & 0.723 & 0.969 \\
    \textcolor{gray}{Specific humidity 0} & \textcolor{gray}{\textbf{0.528}} & \textcolor{gray}{0.594} & \textcolor{gray}{0.708} \\
    \textcolor{gray}{Specific humidity 1} & \textcolor{gray}{\textbf{0.529}} & \textcolor{gray}{0.595} & \textcolor{gray}{0.701} \\
    \textcolor{gray}{Specific humidity 2} & \textcolor{gray}{\textbf{0.534}} & \textcolor{gray}{0.600} & \textcolor{gray}{0.723} \\
    \textcolor{gray}{Specific humidity 3} & \textcolor{gray}{\textbf{0.558}} & \textcolor{gray}{0.621} & \textcolor{gray}{0.764} \\
    \textcolor{gray}{Specific humidity 4} & \textcolor{gray}{\textbf{0.591}} & \textcolor{gray}{0.663} & \textcolor{gray}{0.833} \\
    \textcolor{gray}{Specific humidity 5} & \textcolor{gray}{\textbf{0.644}} & \textcolor{gray}{0.741} & \textcolor{gray}{0.961} \\
    \textcolor{gray}{Specific humidity 6} & \textcolor{gray}{\textbf{0.726}} & \textcolor{gray}{0.785} & \textcolor{gray}{1.09} \\
    \textcolor{gray}{Specific humidity 7} & \textcolor{gray}{\textbf{0.802}} & \textcolor{gray}{0.852} & \textcolor{gray}{1.22} \\
    \textcolor{gray}{Specific humidity 8} & \textcolor{gray}{\textbf{0.837}} & \textcolor{gray}{0.920} & \textcolor{gray}{1.34} \\
    \textcolor{gray}{Specific humidity 9} & \textcolor{gray}{\textbf{0.857}} & \textcolor{gray}{0.865} & \textcolor{gray}{1.35} \\
    Cloud fraction (3D) / 1 & \textbf{0.0376} & 0.0401 & 0.0413 \\
    \textcolor{gray}{Cloud fraction 0} & \textcolor{gray}{\textbf{0.0708}} & \textcolor{gray}{0.0712} & \textcolor{gray}{0.0779} \\
    \textcolor{gray}{Cloud fraction 1} & \textcolor{gray}{\textbf{0.0635}} & \textcolor{gray}{0.0639} & \textcolor{gray}{0.0701} \\
    \textcolor{gray}{Cloud fraction 2} & \textcolor{gray}{\textbf{0.0471}} & \textcolor{gray}{0.0484} & \textcolor{gray}{0.0508} \\
    \textcolor{gray}{Cloud fraction 3} & \textcolor{gray}{\textbf{0.0401}} & \textcolor{gray}{0.0433} & \textcolor{gray}{0.0429} \\
    \textcolor{gray}{Cloud fraction 4} & \textcolor{gray}{\textbf{0.0374}} & \textcolor{gray}{0.0398} & \textcolor{gray}{0.0401} \\
    \textcolor{gray}{Cloud fraction 5} & \textcolor{gray}{\textbf{0.0394}} & \textcolor{gray}{0.0423} & \textcolor{gray}{0.0450} \\
    \textcolor{gray}{Cloud fraction 6} & \textcolor{gray}{\textbf{0.0307}} & \textcolor{gray}{0.0331} & \textcolor{gray}{0.0344} \\
    \textcolor{gray}{Cloud fraction 7} & \textcolor{gray}{\textbf{0.0267}} & \textcolor{gray}{0.0329} & \textcolor{gray}{0.0293} \\
    \textcolor{gray}{Cloud fraction 8} & \textcolor{gray}{\textbf{0.0177}} & \textcolor{gray}{0.0236} & \textcolor{gray}{0.0206} \\
    \textcolor{gray}{Cloud fraction 9} & \textcolor{gray}{0.00251} & \textcolor{gray}{0.00276} & \textcolor{gray}{\textbf{0.00217}} \\
    East--west wind (3D) / m\,s$^{-1}$ & \textbf{6.65} & 7.68 & 7.34 \\
    \textcolor{gray}{East--west wind 0} & \textcolor{gray}{\textbf{2.07}} & \textcolor{gray}{2.49} & \textcolor{gray}{2.21} \\
    \textcolor{gray}{East--west wind 1} & \textcolor{gray}{\textbf{2.57}} & \textcolor{gray}{3.03} & \textcolor{gray}{2.73} \\
    \textcolor{gray}{East--west wind 2} & \textcolor{gray}{\textbf{3.58}} & \textcolor{gray}{4.19} & \textcolor{gray}{3.75} \\
    \textcolor{gray}{East--west wind 3} & \textcolor{gray}{\textbf{4.87}} & \textcolor{gray}{5.74} & \textcolor{gray}{4.99} \\
    \textcolor{gray}{East--west wind 4} & \textcolor{gray}{\textbf{6.06}} & \textcolor{gray}{7.14} & \textcolor{gray}{6.23} \\
    \textcolor{gray}{East--west wind 5} & \textcolor{gray}{\textbf{6.81}} & \textcolor{gray}{8.05} & \textcolor{gray}{7.06} \\
    \textcolor{gray}{East--west wind 6} & \textcolor{gray}{\textbf{7.53}} & \textcolor{gray}{8.80} & \textcolor{gray}{7.98} \\
    \textcolor{gray}{East--west wind 7} & \textcolor{gray}{\textbf{8.33}} & \textcolor{gray}{9.43} & \textcolor{gray}{9.06} \\
    \textcolor{gray}{East--west wind 8} & \textcolor{gray}{\textbf{10.4}} & \textcolor{gray}{12.4} & \textcolor{gray}{11.6} \\
    \textcolor{gray}{East--west wind 9} & \textcolor{gray}{\textbf{14.2}} & \textcolor{gray}{15.5} & \textcolor{gray}{17.7} \\
    North--south wind (3D) / m\,s$^{-1}$ & \textbf{2.91} & 3.45 & 2.91 \\
    \textcolor{gray}{North--south wind 0} & \textcolor{gray}{\textbf{1.42}} & \textcolor{gray}{1.75} & \textcolor{gray}{1.47} \\
    \textcolor{gray}{North--south wind 1} & \textcolor{gray}{\textbf{1.74}} & \textcolor{gray}{2.13} & \textcolor{gray}{1.78} \\
    \textcolor{gray}{North--south wind 2} & \textcolor{gray}{2.43} & \textcolor{gray}{2.92} & \textcolor{gray}{\textbf{2.39}} \\
    \textcolor{gray}{North--south wind 3} & \textcolor{gray}{3.10} & \textcolor{gray}{3.65} & \textcolor{gray}{\textbf{3.09}} \\
    \textcolor{gray}{North--south wind 4} & \textcolor{gray}{3.56} & \textcolor{gray}{4.19} & \textcolor{gray}{\textbf{3.54}} \\
    \textcolor{gray}{North--south wind 5} & \textcolor{gray}{\textbf{3.60}} & \textcolor{gray}{4.39} & \textcolor{gray}{3.62} \\
    \textcolor{gray}{North--south wind 6} & \textcolor{gray}{\textbf{3.59}} & \textcolor{gray}{4.38} & \textcolor{gray}{3.63} \\
    \textcolor{gray}{North--south wind 7} & \textcolor{gray}{3.37} & \textcolor{gray}{4.03} & \textcolor{gray}{\textbf{3.36}} \\
    \textcolor{gray}{North--south wind 8} & \textcolor{gray}{\textbf{3.03}} & \textcolor{gray}{3.68} & \textcolor{gray}{3.03} \\
    \textcolor{gray}{North--south wind 9} & \textcolor{gray}{3.26} & \textcolor{gray}{3.37} & \textcolor{gray}{\textbf{3.23}} \\
    Absorbed shortwave radiation / W\,m$^{-2}$ & \textbf{14.8} & 18.0 & 31.6 \\
    Outgoing longwave radiation / W\,m$^{-2}$ & \textbf{9.76} & 11.4 & 15.5 \\
    \bottomrule
  \end{tabular}
  \caption{\emph{Exoclimate emulation:} Energy score. Bold indicates lower (better) values. Rows 0--9 index pressure levels from surface to top for 3D variables; the individual levels are shown in gray. Scores are averaged over five training seeds.}
  \label{tab:exoplanet-Z}
\end{table*}

\begin{table*}[t]
  \centering
  \scriptsize
  \setlength{\tabcolsep}{5pt}
  \renewcommand{\arraystretch}{1.1}
  \begin{tabular}{@{}p{0.3\textwidth}rrrrrr@{}}
    \toprule
    Variable & GPLFR & PPCA-ICM & SV-LMC & PPCA-MLP & kNN & Train-mean \\
    \midrule
    Surface temperature / K & \textbf{9.32} & 10.4 & 10.3 & 11.5 & 18.3 & 28.0 \\
    Temperature (3D) / K & \textbf{7.76} & 8.73 & 9.66 & 10.2 & 14.0 & 22.2 \\
    \textcolor{gray}{Temperature 0} & \textcolor{gray}{\textbf{8.17}} & \textcolor{gray}{8.85} & \textcolor{gray}{9.18} & \textcolor{gray}{9.50} & \textcolor{gray}{13.6} & \textcolor{gray}{21.4} \\
    \textcolor{gray}{Temperature 1} & \textcolor{gray}{\textbf{8.02}} & \textcolor{gray}{8.78} & \textcolor{gray}{8.96} & \textcolor{gray}{9.35} & \textcolor{gray}{13.4} & \textcolor{gray}{20.3} \\
    \textcolor{gray}{Temperature 2} & \textcolor{gray}{\textbf{7.56}} & \textcolor{gray}{8.46} & \textcolor{gray}{8.51} & \textcolor{gray}{9.05} & \textcolor{gray}{13.0} & \textcolor{gray}{19.4} \\
    \textcolor{gray}{Temperature 3} & \textcolor{gray}{\textbf{6.92}} & \textcolor{gray}{7.78} & \textcolor{gray}{7.94} & \textcolor{gray}{8.64} & \textcolor{gray}{13.0} & \textcolor{gray}{19.5} \\
    \textcolor{gray}{Temperature 4} & \textcolor{gray}{\textbf{6.36}} & \textcolor{gray}{7.13} & \textcolor{gray}{7.52} & \textcolor{gray}{8.02} & \textcolor{gray}{13.3} & \textcolor{gray}{19.8} \\
    \textcolor{gray}{Temperature 5} & \textcolor{gray}{\textbf{6.20}} & \textcolor{gray}{7.24} & \textcolor{gray}{7.97} & \textcolor{gray}{8.28} & \textcolor{gray}{13.2} & \textcolor{gray}{20.5} \\
    \textcolor{gray}{Temperature 6} & \textcolor{gray}{\textbf{6.60}} & \textcolor{gray}{7.58} & \textcolor{gray}{9.14} & \textcolor{gray}{9.27} & \textcolor{gray}{13.8} & \textcolor{gray}{21.6} \\
    \textcolor{gray}{Temperature 7} & \textcolor{gray}{\textbf{9.02}} & \textcolor{gray}{9.75} & \textcolor{gray}{11.7} & \textcolor{gray}{10.6} & \textcolor{gray}{15.5} & \textcolor{gray}{25.2} \\
    \textcolor{gray}{Temperature 8} & \textcolor{gray}{\textbf{10.7}} & \textcolor{gray}{12.1} & \textcolor{gray}{13.7} & \textcolor{gray}{12.3} & \textcolor{gray}{16.9} & \textcolor{gray}{27.1} \\
    \textcolor{gray}{Temperature 9} & \textcolor{gray}{\textbf{8.04}} & \textcolor{gray}{9.59} & \textcolor{gray}{12.0} & \textcolor{gray}{17.0} & \textcolor{gray}{14.4} & \textcolor{gray}{26.7} \\
    Specific humidity (3D) / dex & \textbf{0.379} & 0.402 & 0.469 & 0.518 & 0.720 & 1.16 \\
    \textcolor{gray}{Specific humidity 0} & \textcolor{gray}{\textbf{0.300}} & \textcolor{gray}{0.329} & \textcolor{gray}{0.346} & \textcolor{gray}{0.383} & \textcolor{gray}{0.550} & \textcolor{gray}{0.789} \\
    \textcolor{gray}{Specific humidity 1} & \textcolor{gray}{\textbf{0.299}} & \textcolor{gray}{0.329} & \textcolor{gray}{0.346} & \textcolor{gray}{0.379} & \textcolor{gray}{0.558} & \textcolor{gray}{0.805} \\
    \textcolor{gray}{Specific humidity 2} & \textcolor{gray}{\textbf{0.304}} & \textcolor{gray}{0.337} & \textcolor{gray}{0.350} & \textcolor{gray}{0.380} & \textcolor{gray}{0.575} & \textcolor{gray}{0.863} \\
    \textcolor{gray}{Specific humidity 3} & \textcolor{gray}{\textbf{0.321}} & \textcolor{gray}{0.350} & \textcolor{gray}{0.370} & \textcolor{gray}{0.403} & \textcolor{gray}{0.613} & \textcolor{gray}{0.940} \\
    \textcolor{gray}{Specific humidity 4} & \textcolor{gray}{\textbf{0.343}} & \textcolor{gray}{0.372} & \textcolor{gray}{0.406} & \textcolor{gray}{0.446} & \textcolor{gray}{0.668} & \textcolor{gray}{1.04} \\
    \textcolor{gray}{Specific humidity 5} & \textcolor{gray}{\textbf{0.371}} & \textcolor{gray}{0.414} & \textcolor{gray}{0.464} & \textcolor{gray}{0.508} & \textcolor{gray}{0.765} & \textcolor{gray}{1.20} \\
    \textcolor{gray}{Specific humidity 6} & \textcolor{gray}{\textbf{0.412}} & \textcolor{gray}{0.434} & \textcolor{gray}{0.522} & \textcolor{gray}{0.548} & \textcolor{gray}{0.873} & \textcolor{gray}{1.40} \\
    \textcolor{gray}{Specific humidity 7} & \textcolor{gray}{\textbf{0.455}} & \textcolor{gray}{0.472} & \textcolor{gray}{0.582} & \textcolor{gray}{0.584} & \textcolor{gray}{0.923} & \textcolor{gray}{1.57} \\
    \textcolor{gray}{Specific humidity 8} & \textcolor{gray}{\textbf{0.479}} & \textcolor{gray}{0.501} & \textcolor{gray}{0.618} & \textcolor{gray}{0.594} & \textcolor{gray}{0.883} & \textcolor{gray}{1.57} \\
    \textcolor{gray}{Specific humidity 9} & \textcolor{gray}{0.509} & \textcolor{gray}{\textbf{0.487}} & \textcolor{gray}{0.683} & \textcolor{gray}{0.958} & \textcolor{gray}{0.787} & \textcolor{gray}{1.46} \\
    Cloud fraction (3D) / 1 & \textbf{0.0504} & 0.0529 & 0.0569 & 0.0591 & 0.0575 & 0.110 \\
    \textcolor{gray}{Cloud fraction 0} & \textcolor{gray}{\textbf{0.0970}} & \textcolor{gray}{0.0982} & \textcolor{gray}{0.108} & \textcolor{gray}{0.113} & \textcolor{gray}{0.111} & \textcolor{gray}{0.247} \\
    \textcolor{gray}{Cloud fraction 1} & \textcolor{gray}{\textbf{0.0863}} & \textcolor{gray}{0.0864} & \textcolor{gray}{0.0971} & \textcolor{gray}{0.0996} & \textcolor{gray}{0.0969} & \textcolor{gray}{0.206} \\
    \textcolor{gray}{Cloud fraction 2} & \textcolor{gray}{\textbf{0.0639}} & \textcolor{gray}{0.0654} & \textcolor{gray}{0.0713} & \textcolor{gray}{0.0742} & \textcolor{gray}{0.0723} & \textcolor{gray}{0.134} \\
    \textcolor{gray}{Cloud fraction 3} & \textcolor{gray}{\textbf{0.0544}} & \textcolor{gray}{0.0577} & \textcolor{gray}{0.0600} & \textcolor{gray}{0.0638} & \textcolor{gray}{0.0591} & \textcolor{gray}{0.110} \\
    \textcolor{gray}{Cloud fraction 4} & \textcolor{gray}{\textbf{0.0496}} & \textcolor{gray}{0.0520} & \textcolor{gray}{0.0547} & \textcolor{gray}{0.0600} & \textcolor{gray}{0.0542} & \textcolor{gray}{0.0994} \\
    \textcolor{gray}{Cloud fraction 5} & \textcolor{gray}{\textbf{0.0514}} & \textcolor{gray}{0.0532} & \textcolor{gray}{0.0596} & \textcolor{gray}{0.0628} & \textcolor{gray}{0.0571} & \textcolor{gray}{0.0989} \\
    \textcolor{gray}{Cloud fraction 6} & \textcolor{gray}{\textbf{0.0396}} & \textcolor{gray}{0.0421} & \textcolor{gray}{0.0470} & \textcolor{gray}{0.0480} & \textcolor{gray}{0.0475} & \textcolor{gray}{0.0831} \\
    \textcolor{gray}{Cloud fraction 7} & \textcolor{gray}{\textbf{0.0349}} & \textcolor{gray}{0.0410} & \textcolor{gray}{0.0417} & \textcolor{gray}{0.0404} & \textcolor{gray}{0.0458} & \textcolor{gray}{0.0665} \\
    \textcolor{gray}{Cloud fraction 8} & \textcolor{gray}{\textbf{0.0245}} & \textcolor{gray}{0.0298} & \textcolor{gray}{0.0274} & \textcolor{gray}{0.0252} & \textcolor{gray}{0.0284} & \textcolor{gray}{0.0395} \\
    \textcolor{gray}{Cloud fraction 9} & \textcolor{gray}{\textbf{0.00199}} & \textcolor{gray}{0.00356} & \textcolor{gray}{0.00217} & \textcolor{gray}{0.00384} & \textcolor{gray}{0.00265} & \textcolor{gray}{0.0100} \\
    East--west wind (3D) / m\,s$^{-1}$ & \textbf{8.51} & 9.63 & 9.67 & 10.4 & 10.1 & 17.1 \\
    \textcolor{gray}{East--west wind 0} & \textcolor{gray}{\textbf{2.77}} & \textcolor{gray}{3.19} & \textcolor{gray}{2.97} & \textcolor{gray}{3.20} & \textcolor{gray}{3.37} & \textcolor{gray}{4.45} \\
    \textcolor{gray}{East--west wind 1} & \textcolor{gray}{\textbf{3.38}} & \textcolor{gray}{3.84} & \textcolor{gray}{3.65} & \textcolor{gray}{3.88} & \textcolor{gray}{4.19} & \textcolor{gray}{5.68} \\
    \textcolor{gray}{East--west wind 2} & \textcolor{gray}{\textbf{4.63}} & \textcolor{gray}{5.28} & \textcolor{gray}{5.00} & \textcolor{gray}{5.32} & \textcolor{gray}{5.95} & \textcolor{gray}{8.72} \\
    \textcolor{gray}{East--west wind 3} & \textcolor{gray}{\textbf{6.25}} & \textcolor{gray}{7.19} & \textcolor{gray}{6.70} & \textcolor{gray}{7.10} & \textcolor{gray}{8.01} & \textcolor{gray}{12.9} \\
    \textcolor{gray}{East--west wind 4} & \textcolor{gray}{\textbf{7.76}} & \textcolor{gray}{8.94} & \textcolor{gray}{8.34} & \textcolor{gray}{8.84} & \textcolor{gray}{9.88} & \textcolor{gray}{17.2} \\
    \textcolor{gray}{East--west wind 5} & \textcolor{gray}{\textbf{8.72}} & \textcolor{gray}{10.2} & \textcolor{gray}{9.41} & \textcolor{gray}{10.0} & \textcolor{gray}{11.4} & \textcolor{gray}{20.2} \\
    \textcolor{gray}{East--west wind 6} & \textcolor{gray}{\textbf{9.70}} & \textcolor{gray}{11.2} & \textcolor{gray}{10.6} & \textcolor{gray}{11.1} & \textcolor{gray}{12.6} & \textcolor{gray}{22.2} \\
    \textcolor{gray}{East--west wind 7} & \textcolor{gray}{\textbf{10.6}} & \textcolor{gray}{11.9} & \textcolor{gray}{12.0} & \textcolor{gray}{12.3} & \textcolor{gray}{12.3} & \textcolor{gray}{23.5} \\
    \textcolor{gray}{East--west wind 8} & \textcolor{gray}{\textbf{13.1}} & \textcolor{gray}{15.2} & \textcolor{gray}{15.1} & \textcolor{gray}{15.6} & \textcolor{gray}{13.8} & \textcolor{gray}{25.4} \\
    \textcolor{gray}{East--west wind 9} & \textcolor{gray}{\textbf{18.2}} & \textcolor{gray}{19.4} & \textcolor{gray}{23.0} & \textcolor{gray}{26.7} & \textcolor{gray}{19.1} & \textcolor{gray}{31.3} \\
    North--south wind (3D) / m\,s$^{-1}$ & \textbf{3.70} & 4.31 & 3.88 & 4.47 & 4.34 & 6.39 \\
    \textcolor{gray}{North--south wind 0} & \textcolor{gray}{\textbf{1.90}} & \textcolor{gray}{2.22} & \textcolor{gray}{2.00} & \textcolor{gray}{2.22} & \textcolor{gray}{2.28} & \textcolor{gray}{2.89} \\
    \textcolor{gray}{North--south wind 1} & \textcolor{gray}{\textbf{2.31}} & \textcolor{gray}{2.68} & \textcolor{gray}{2.39} & \textcolor{gray}{2.65} & \textcolor{gray}{2.74} & \textcolor{gray}{3.52} \\
    \textcolor{gray}{North--south wind 2} & \textcolor{gray}{\textbf{3.17}} & \textcolor{gray}{3.63} & \textcolor{gray}{3.24} & \textcolor{gray}{3.56} & \textcolor{gray}{3.68} & \textcolor{gray}{4.90} \\
    \textcolor{gray}{North--south wind 3} & \textcolor{gray}{\textbf{3.98}} & \textcolor{gray}{4.51} & \textcolor{gray}{4.13} & \textcolor{gray}{4.52} & \textcolor{gray}{4.63} & \textcolor{gray}{6.36} \\
    \textcolor{gray}{North--south wind 4} & \textcolor{gray}{\textbf{4.59}} & \textcolor{gray}{5.20} & \textcolor{gray}{4.78} & \textcolor{gray}{5.26} & \textcolor{gray}{5.38} & \textcolor{gray}{7.35} \\
    \textcolor{gray}{North--south wind 5} & \textcolor{gray}{\textbf{4.67}} & \textcolor{gray}{5.49} & \textcolor{gray}{4.88} & \textcolor{gray}{5.45} & \textcolor{gray}{5.66} & \textcolor{gray}{7.83} \\
    \textcolor{gray}{North--south wind 6} & \textcolor{gray}{\textbf{4.65}} & \textcolor{gray}{5.50} & \textcolor{gray}{4.93} & \textcolor{gray}{5.36} & \textcolor{gray}{5.79} & \textcolor{gray}{8.04} \\
    \textcolor{gray}{North--south wind 7} & \textcolor{gray}{\textbf{4.31}} & \textcolor{gray}{5.08} & \textcolor{gray}{4.43} & \textcolor{gray}{4.94} & \textcolor{gray}{5.29} & \textcolor{gray}{7.57} \\
    \textcolor{gray}{North--south wind 8} & \textcolor{gray}{\textbf{3.67}} & \textcolor{gray}{4.64} & \textcolor{gray}{3.88} & \textcolor{gray}{4.57} & \textcolor{gray}{4.20} & \textcolor{gray}{6.97} \\
    \textcolor{gray}{North--south wind 9} & \textcolor{gray}{3.75} & \textcolor{gray}{4.15} & \textcolor{gray}{4.13} & \textcolor{gray}{6.17} & \textcolor{gray}{\textbf{3.72}} & \textcolor{gray}{8.48} \\
    Absorbed shortwave radiation / W\,m$^{-2}$ & \textbf{19.6} & 23.5 & 22.5 & 28.7 & 47.7 & 343 \\
    Outgoing longwave radiation / W\,m$^{-2}$ & \textbf{12.8} & 15.3 & 16.2 & 16.5 & 28.9 & 48.2 \\
    \bottomrule
  \end{tabular}
  \caption{\emph{Exoclimate emulation:} RMSE. Bold indicates lower (better) values. Rows 0--9 index pressure levels from surface to top for 3D variables; the individual levels are shown in gray. Scores for learned methods are the mean over five training seeds; kNN and the training mean are deterministic.}
  \label{tab:exoplanet-AA}
\end{table*}

\begin{table*}[t]
  \centering
  \scriptsize
  \setlength{\tabcolsep}{5pt}
  \renewcommand{\arraystretch}{1.1}
  \begin{tabular}{@{}p{0.3\textwidth}rrrrrr@{}}
    \toprule
    Variable & GPLFR & PPCA-ICM & SV-LMC & PPCA-MLP & kNN & Train-mean \\
    \midrule
    Surface temperature & \textbf{0.978} & 0.973 & 0.974 & 0.968 & 0.971 & 0.939 \\
    Temperature (3D) & \textbf{0.688} & 0.629 & 0.632 & 0.614 & 0.639 & 0.380 \\
    \textcolor{gray}{Temperature 0} & \textcolor{gray}{\textbf{0.895}} & \textcolor{gray}{0.876} & \textcolor{gray}{0.882} & \textcolor{gray}{0.880} & \textcolor{gray}{0.887} & \textcolor{gray}{0.842} \\
    \textcolor{gray}{Temperature 1} & \textcolor{gray}{\textbf{0.875}} & \textcolor{gray}{0.851} & \textcolor{gray}{0.861} & \textcolor{gray}{0.854} & \textcolor{gray}{0.868} & \textcolor{gray}{0.797} \\
    \textcolor{gray}{Temperature 2} & \textcolor{gray}{\textbf{0.843}} & \textcolor{gray}{0.775} & \textcolor{gray}{0.804} & \textcolor{gray}{0.794} & \textcolor{gray}{0.833} & \textcolor{gray}{0.747} \\
    \textcolor{gray}{Temperature 3} & \textcolor{gray}{\textbf{0.778}} & \textcolor{gray}{0.716} & \textcolor{gray}{0.725} & \textcolor{gray}{0.701} & \textcolor{gray}{0.752} & \textcolor{gray}{0.629} \\
    \textcolor{gray}{Temperature 4} & \textcolor{gray}{\textbf{0.709}} & \textcolor{gray}{0.621} & \textcolor{gray}{0.655} & \textcolor{gray}{0.651} & \textcolor{gray}{0.592} & \textcolor{gray}{0.425} \\
    \textcolor{gray}{Temperature 5} & \textcolor{gray}{\textbf{0.621}} & \textcolor{gray}{0.538} & \textcolor{gray}{0.544} & \textcolor{gray}{0.538} & \textcolor{gray}{0.543} & \textcolor{gray}{0.177} \\
    \textcolor{gray}{Temperature 6} & \textcolor{gray}{\textbf{0.612}} & \textcolor{gray}{0.510} & \textcolor{gray}{0.505} & \textcolor{gray}{0.462} & \textcolor{gray}{0.558} & \textcolor{gray}{-0.0532} \\
    \textcolor{gray}{Temperature 7} & \textcolor{gray}{\textbf{0.601}} & \textcolor{gray}{0.498} & \textcolor{gray}{0.530} & \textcolor{gray}{0.525} & \textcolor{gray}{0.501} & \textcolor{gray}{0.00427} \\
    \textcolor{gray}{Temperature 8} & \textcolor{gray}{0.413} & \textcolor{gray}{\textbf{0.417}} & \textcolor{gray}{0.387} & \textcolor{gray}{0.349} & \textcolor{gray}{0.390} & \textcolor{gray}{0.118} \\
    \textcolor{gray}{Temperature 9} & \textcolor{gray}{\textbf{0.533}} & \textcolor{gray}{0.486} & \textcolor{gray}{0.425} & \textcolor{gray}{0.390} & \textcolor{gray}{0.471} & \textcolor{gray}{0.111} \\
    Specific humidity (3D) & \textbf{0.819} & 0.782 & 0.796 & 0.781 & 0.810 & 0.687 \\
    \textcolor{gray}{Specific humidity 0} & \textcolor{gray}{0.936} & \textcolor{gray}{0.923} & \textcolor{gray}{\textbf{0.936}} & \textcolor{gray}{0.926} & \textcolor{gray}{0.922} & \textcolor{gray}{0.882} \\
    \textcolor{gray}{Specific humidity 1} & \textcolor{gray}{\textbf{0.926}} & \textcolor{gray}{0.909} & \textcolor{gray}{0.925} & \textcolor{gray}{0.915} & \textcolor{gray}{0.903} & \textcolor{gray}{0.844} \\
    \textcolor{gray}{Specific humidity 2} & \textcolor{gray}{\textbf{0.909}} & \textcolor{gray}{0.881} & \textcolor{gray}{0.904} & \textcolor{gray}{0.888} & \textcolor{gray}{0.879} & \textcolor{gray}{0.787} \\
    \textcolor{gray}{Specific humidity 3} & \textcolor{gray}{\textbf{0.892}} & \textcolor{gray}{0.864} & \textcolor{gray}{0.886} & \textcolor{gray}{0.868} & \textcolor{gray}{0.865} & \textcolor{gray}{0.734} \\
    \textcolor{gray}{Specific humidity 4} & \textcolor{gray}{\textbf{0.898}} & \textcolor{gray}{0.874} & \textcolor{gray}{0.891} & \textcolor{gray}{0.877} & \textcolor{gray}{0.865} & \textcolor{gray}{0.712} \\
    \textcolor{gray}{Specific humidity 5} & \textcolor{gray}{\textbf{0.900}} & \textcolor{gray}{0.872} & \textcolor{gray}{0.895} & \textcolor{gray}{0.880} & \textcolor{gray}{0.862} & \textcolor{gray}{0.720} \\
    \textcolor{gray}{Specific humidity 6} & \textcolor{gray}{\textbf{0.876}} & \textcolor{gray}{0.835} & \textcolor{gray}{0.850} & \textcolor{gray}{0.829} & \textcolor{gray}{0.834} & \textcolor{gray}{0.710} \\
    \textcolor{gray}{Specific humidity 7} & \textcolor{gray}{\textbf{0.815}} & \textcolor{gray}{0.786} & \textcolor{gray}{0.762} & \textcolor{gray}{0.767} & \textcolor{gray}{0.792} & \textcolor{gray}{0.677} \\
    \textcolor{gray}{Specific humidity 8} & \textcolor{gray}{0.605} & \textcolor{gray}{0.567} & \textcolor{gray}{0.544} & \textcolor{gray}{0.536} & \textcolor{gray}{\textbf{0.632}} & \textcolor{gray}{0.540} \\
    \textcolor{gray}{Specific humidity 9} & \textcolor{gray}{0.435} & \textcolor{gray}{0.310} & \textcolor{gray}{0.366} & \textcolor{gray}{0.328} & \textcolor{gray}{\textbf{0.546}} & \textcolor{gray}{0.260} \\
    Cloud fraction (3D) & \textbf{0.667} & 0.649 & 0.640 & 0.620 & 0.666 & 0.486 \\
    \textcolor{gray}{Cloud fraction 0} & \textcolor{gray}{\textbf{0.734}} & \textcolor{gray}{0.692} & \textcolor{gray}{0.693} & \textcolor{gray}{0.662} & \textcolor{gray}{0.688} & \textcolor{gray}{0.350} \\
    \textcolor{gray}{Cloud fraction 1} & \textcolor{gray}{\textbf{0.734}} & \textcolor{gray}{0.718} & \textcolor{gray}{0.699} & \textcolor{gray}{0.664} & \textcolor{gray}{0.710} & \textcolor{gray}{0.396} \\
    \textcolor{gray}{Cloud fraction 2} & \textcolor{gray}{0.771} & \textcolor{gray}{0.769} & \textcolor{gray}{0.730} & \textcolor{gray}{0.692} & \textcolor{gray}{\textbf{0.785}} & \textcolor{gray}{0.564} \\
    \textcolor{gray}{Cloud fraction 3} & \textcolor{gray}{\textbf{0.809}} & \textcolor{gray}{0.793} & \textcolor{gray}{0.779} & \textcolor{gray}{0.762} & \textcolor{gray}{0.808} & \textcolor{gray}{0.658} \\
    \textcolor{gray}{Cloud fraction 4} & \textcolor{gray}{0.787} & \textcolor{gray}{0.764} & \textcolor{gray}{0.760} & \textcolor{gray}{0.748} & \textcolor{gray}{\textbf{0.796}} & \textcolor{gray}{0.636} \\
    \textcolor{gray}{Cloud fraction 5} & \textcolor{gray}{0.727} & \textcolor{gray}{0.722} & \textcolor{gray}{0.707} & \textcolor{gray}{0.699} & \textcolor{gray}{\textbf{0.745}} & \textcolor{gray}{0.604} \\
    \textcolor{gray}{Cloud fraction 6} & \textcolor{gray}{0.695} & \textcolor{gray}{0.679} & \textcolor{gray}{0.674} & \textcolor{gray}{0.675} & \textcolor{gray}{\textbf{0.713}} & \textcolor{gray}{0.549} \\
    \textcolor{gray}{Cloud fraction 7} & \textcolor{gray}{0.671} & \textcolor{gray}{0.648} & \textcolor{gray}{0.655} & \textcolor{gray}{0.651} & \textcolor{gray}{\textbf{0.676}} & \textcolor{gray}{0.531} \\
    \textcolor{gray}{Cloud fraction 8} & \textcolor{gray}{\textbf{0.511}} & \textcolor{gray}{0.481} & \textcolor{gray}{0.492} & \textcolor{gray}{0.471} & \textcolor{gray}{0.499} & \textcolor{gray}{0.427} \\
    \textcolor{gray}{Cloud fraction 9} & \textcolor{gray}{0.234} & \textcolor{gray}{0.227} & \textcolor{gray}{0.207} & \textcolor{gray}{0.179} & \textcolor{gray}{\textbf{0.236}} & \textcolor{gray}{0.144} \\
    East--west wind (3D) & \textbf{0.712} & 0.666 & 0.680 & 0.653 & 0.677 & 0.508 \\
    \textcolor{gray}{East--west wind 0} & \textcolor{gray}{\textbf{0.850}} & \textcolor{gray}{0.808} & \textcolor{gray}{0.840} & \textcolor{gray}{0.832} & \textcolor{gray}{0.807} & \textcolor{gray}{0.721} \\
    \textcolor{gray}{East--west wind 1} & \textcolor{gray}{\textbf{0.799}} & \textcolor{gray}{0.751} & \textcolor{gray}{0.791} & \textcolor{gray}{0.778} & \textcolor{gray}{0.760} & \textcolor{gray}{0.633} \\
    \textcolor{gray}{East--west wind 2} & \textcolor{gray}{0.722} & \textcolor{gray}{0.675} & \textcolor{gray}{\textbf{0.725}} & \textcolor{gray}{0.691} & \textcolor{gray}{0.696} & \textcolor{gray}{0.503} \\
    \textcolor{gray}{East--west wind 3} & \textcolor{gray}{\textbf{0.736}} & \textcolor{gray}{0.671} & \textcolor{gray}{0.714} & \textcolor{gray}{0.685} & \textcolor{gray}{0.692} & \textcolor{gray}{0.513} \\
    \textcolor{gray}{East--west wind 4} & \textcolor{gray}{\textbf{0.761}} & \textcolor{gray}{0.706} & \textcolor{gray}{0.733} & \textcolor{gray}{0.701} & \textcolor{gray}{0.702} & \textcolor{gray}{0.533} \\
    \textcolor{gray}{East--west wind 5} & \textcolor{gray}{\textbf{0.763}} & \textcolor{gray}{0.706} & \textcolor{gray}{0.742} & \textcolor{gray}{0.704} & \textcolor{gray}{0.727} & \textcolor{gray}{0.542} \\
    \textcolor{gray}{East--west wind 6} & \textcolor{gray}{\textbf{0.758}} & \textcolor{gray}{0.697} & \textcolor{gray}{0.731} & \textcolor{gray}{0.699} & \textcolor{gray}{0.710} & \textcolor{gray}{0.520} \\
    \textcolor{gray}{East--west wind 7} & \textcolor{gray}{\textbf{0.659}} & \textcolor{gray}{0.636} & \textcolor{gray}{0.606} & \textcolor{gray}{0.610} & \textcolor{gray}{0.632} & \textcolor{gray}{0.453} \\
    \textcolor{gray}{East--west wind 8} & \textcolor{gray}{\textbf{0.599}} & \textcolor{gray}{0.532} & \textcolor{gray}{0.524} & \textcolor{gray}{0.497} & \textcolor{gray}{0.557} & \textcolor{gray}{0.344} \\
    \textcolor{gray}{East--west wind 9} & \textcolor{gray}{0.473} & \textcolor{gray}{0.480} & \textcolor{gray}{0.395} & \textcolor{gray}{0.329} & \textcolor{gray}{\textbf{0.489}} & \textcolor{gray}{0.312} \\
    North--south wind (3D) & \textbf{0.690} & 0.644 & 0.666 & 0.635 & 0.683 & 0.516 \\
    \textcolor{gray}{North--south wind 0} & \textcolor{gray}{\textbf{0.796}} & \textcolor{gray}{0.742} & \textcolor{gray}{0.770} & \textcolor{gray}{0.741} & \textcolor{gray}{0.737} & \textcolor{gray}{0.614} \\
    \textcolor{gray}{North--south wind 1} & \textcolor{gray}{\textbf{0.767}} & \textcolor{gray}{0.704} & \textcolor{gray}{0.739} & \textcolor{gray}{0.713} & \textcolor{gray}{0.704} & \textcolor{gray}{0.527} \\
    \textcolor{gray}{North--south wind 2} & \textcolor{gray}{\textbf{0.691}} & \textcolor{gray}{0.630} & \textcolor{gray}{0.681} & \textcolor{gray}{0.661} & \textcolor{gray}{0.676} & \textcolor{gray}{0.421} \\
    \textcolor{gray}{North--south wind 3} & \textcolor{gray}{\textbf{0.705}} & \textcolor{gray}{0.645} & \textcolor{gray}{0.684} & \textcolor{gray}{0.664} & \textcolor{gray}{0.703} & \textcolor{gray}{0.462} \\
    \textcolor{gray}{North--south wind 4} & \textcolor{gray}{\textbf{0.734}} & \textcolor{gray}{0.686} & \textcolor{gray}{0.707} & \textcolor{gray}{0.690} & \textcolor{gray}{0.722} & \textcolor{gray}{0.543} \\
    \textcolor{gray}{North--south wind 5} & \textcolor{gray}{\textbf{0.740}} & \textcolor{gray}{0.705} & \textcolor{gray}{0.707} & \textcolor{gray}{0.680} & \textcolor{gray}{0.718} & \textcolor{gray}{0.577} \\
    \textcolor{gray}{North--south wind 6} & \textcolor{gray}{\textbf{0.744}} & \textcolor{gray}{0.731} & \textcolor{gray}{0.697} & \textcolor{gray}{0.673} & \textcolor{gray}{0.724} & \textcolor{gray}{0.582} \\
    \textcolor{gray}{North--south wind 7} & \textcolor{gray}{\textbf{0.671}} & \textcolor{gray}{0.636} & \textcolor{gray}{0.624} & \textcolor{gray}{0.614} & \textcolor{gray}{0.637} & \textcolor{gray}{0.564} \\
    \textcolor{gray}{North--south wind 8} & \textcolor{gray}{0.530} & \textcolor{gray}{0.456} & \textcolor{gray}{0.566} & \textcolor{gray}{0.506} & \textcolor{gray}{\textbf{0.652}} & \textcolor{gray}{0.505} \\
    \textcolor{gray}{North--south wind 9} & \textcolor{gray}{0.520} & \textcolor{gray}{0.508} & \textcolor{gray}{0.485} & \textcolor{gray}{0.410} & \textcolor{gray}{\textbf{0.560}} & \textcolor{gray}{0.365} \\
    Absorbed shortwave radiation & \textbf{0.997} & 0.996 & 0.997 & 0.996 & 0.996 & 0.993 \\
    Outgoing longwave radiation & \textbf{0.909} & 0.890 & 0.887 & 0.885 & 0.854 & 0.706 \\
    \bottomrule
  \end{tabular}
  \caption{\emph{Exoclimate emulation:} Anomaly correlation coefficient (ACC). Bold indicates higher (better) values. Rows 0--9 index pressure levels from surface to top for 3D variables; the individual levels are shown in gray. Scores for learned methods are the mean over five training seeds; kNN and the training mean are deterministic.}
  \label{tab:exoplanet-AB}
\end{table*}

\begin{table*}[t]
  \centering
  \scriptsize
  \setlength{\tabcolsep}{5pt}
  \renewcommand{\arraystretch}{1.1}
  \begin{tabular}{@{}p{0.3\textwidth}rrr@{}}
    \toprule
    Variable & GPLFR & PPCA-ICM & SV-LMC \\
    \midrule
    Surface temperature & \textbf{0.571} & 0.536 & 1.57 \\
    Temperature (3D) & \textbf{0.607} & 0.547 & 1.44 \\
    \textcolor{gray}{Temperature 0} & \textcolor{gray}{0.557} & \textcolor{gray}{0.516} & \textcolor{gray}{\textbf{1.43}} \\
    \textcolor{gray}{Temperature 1} & \textcolor{gray}{0.508} & \textcolor{gray}{0.516} & \textcolor{gray}{\textbf{1.40}} \\
    \textcolor{gray}{Temperature 2} & \textcolor{gray}{0.541} & \textcolor{gray}{\textbf{0.542}} & \textcolor{gray}{1.47} \\
    \textcolor{gray}{Temperature 3} & \textcolor{gray}{0.552} & \textcolor{gray}{\textbf{0.581}} & \textcolor{gray}{1.55} \\
    \textcolor{gray}{Temperature 4} & \textcolor{gray}{0.609} & \textcolor{gray}{\textbf{0.653}} & \textcolor{gray}{1.68} \\
    \textcolor{gray}{Temperature 5} & \textcolor{gray}{\textbf{0.691}} & \textcolor{gray}{0.687} & \textcolor{gray}{1.71} \\
    \textcolor{gray}{Temperature 6} & \textcolor{gray}{\textbf{0.805}} & \textcolor{gray}{0.659} & \textcolor{gray}{1.61} \\
    \textcolor{gray}{Temperature 7} & \textcolor{gray}{0.494} & \textcolor{gray}{0.434} & \textcolor{gray}{\textbf{1.26}} \\
    \textcolor{gray}{Temperature 8} & \textcolor{gray}{0.412} & \textcolor{gray}{0.348} & \textcolor{gray}{\textbf{1.06}} \\
    \textcolor{gray}{Temperature 9} & \textcolor{gray}{\textbf{0.903}} & \textcolor{gray}{0.537} & \textcolor{gray}{1.21} \\
    Specific humidity (3D) & \textbf{0.661} & 0.599 & 1.57 \\
    \textcolor{gray}{Specific humidity 0} & \textcolor{gray}{\textbf{0.645}} & \textcolor{gray}{0.528} & \textcolor{gray}{1.50} \\
    \textcolor{gray}{Specific humidity 1} & \textcolor{gray}{\textbf{0.601}} & \textcolor{gray}{0.522} & \textcolor{gray}{1.50} \\
    \textcolor{gray}{Specific humidity 2} & \textcolor{gray}{\textbf{0.633}} & \textcolor{gray}{0.549} & \textcolor{gray}{1.54} \\
    \textcolor{gray}{Specific humidity 3} & \textcolor{gray}{\textbf{0.671}} & \textcolor{gray}{0.598} & \textcolor{gray}{1.64} \\
    \textcolor{gray}{Specific humidity 4} & \textcolor{gray}{\textbf{0.674}} & \textcolor{gray}{0.615} & \textcolor{gray}{1.67} \\
    \textcolor{gray}{Specific humidity 5} & \textcolor{gray}{\textbf{0.666}} & \textcolor{gray}{0.618} & \textcolor{gray}{1.63} \\
    \textcolor{gray}{Specific humidity 6} & \textcolor{gray}{0.605} & \textcolor{gray}{\textbf{0.614}} & \textcolor{gray}{1.58} \\
    \textcolor{gray}{Specific humidity 7} & \textcolor{gray}{0.596} & \textcolor{gray}{\textbf{0.624}} & \textcolor{gray}{1.57} \\
    \textcolor{gray}{Specific humidity 8} & \textcolor{gray}{\textbf{0.675}} & \textcolor{gray}{0.644} & \textcolor{gray}{1.63} \\
    \textcolor{gray}{Specific humidity 9} & \textcolor{gray}{\textbf{0.840}} & \textcolor{gray}{0.677} & \textcolor{gray}{1.42} \\
    Cloud fraction (3D) & \textbf{0.963} & 0.634 & 0.911 \\
    \textcolor{gray}{Cloud fraction 0} & \textcolor{gray}{\textbf{0.989}} & \textcolor{gray}{0.760} & \textcolor{gray}{1.03} \\
    \textcolor{gray}{Cloud fraction 1} & \textcolor{gray}{0.912} & \textcolor{gray}{0.712} & \textcolor{gray}{\textbf{0.999}} \\
    \textcolor{gray}{Cloud fraction 2} & \textcolor{gray}{0.928} & \textcolor{gray}{0.707} & \textcolor{gray}{\textbf{1.02}} \\
    \textcolor{gray}{Cloud fraction 3} & \textcolor{gray}{0.835} & \textcolor{gray}{0.605} & \textcolor{gray}{\textbf{0.909}} \\
    \textcolor{gray}{Cloud fraction 4} & \textcolor{gray}{\textbf{0.830}} & \textcolor{gray}{0.568} & \textcolor{gray}{0.825} \\
    \textcolor{gray}{Cloud fraction 5} & \textcolor{gray}{\textbf{0.805}} & \textcolor{gray}{0.480} & \textcolor{gray}{0.741} \\
    \textcolor{gray}{Cloud fraction 6} & \textcolor{gray}{\textbf{0.980}} & \textcolor{gray}{0.550} & \textcolor{gray}{0.910} \\
    \textcolor{gray}{Cloud fraction 7} & \textcolor{gray}{\textbf{0.898}} & \textcolor{gray}{0.505} & \textcolor{gray}{0.876} \\
    \textcolor{gray}{Cloud fraction 8} & \textcolor{gray}{\textbf{0.826}} & \textcolor{gray}{0.450} & \textcolor{gray}{0.735} \\
    \textcolor{gray}{Cloud fraction 9} & \textcolor{gray}{1.63} & \textcolor{gray}{\textbf{0.999}} & \textcolor{gray}{1.07} \\
    East--west wind (3D) & 0.586 & 0.467 & \textbf{0.717} \\
    \textcolor{gray}{East--west wind 0} & \textcolor{gray}{0.734} & \textcolor{gray}{0.492} & \textcolor{gray}{\textbf{0.802}} \\
    \textcolor{gray}{East--west wind 1} & \textcolor{gray}{0.617} & \textcolor{gray}{0.448} & \textcolor{gray}{\textbf{0.722}} \\
    \textcolor{gray}{East--west wind 2} & \textcolor{gray}{0.541} & \textcolor{gray}{0.422} & \textcolor{gray}{\textbf{0.681}} \\
    \textcolor{gray}{East--west wind 3} & \textcolor{gray}{0.503} & \textcolor{gray}{0.403} & \textcolor{gray}{\textbf{0.663}} \\
    \textcolor{gray}{East--west wind 4} & \textcolor{gray}{0.501} & \textcolor{gray}{0.413} & \textcolor{gray}{\textbf{0.678}} \\
    \textcolor{gray}{East--west wind 5} & \textcolor{gray}{0.507} & \textcolor{gray}{0.436} & \textcolor{gray}{\textbf{0.716}} \\
    \textcolor{gray}{East--west wind 6} & \textcolor{gray}{0.529} & \textcolor{gray}{0.463} & \textcolor{gray}{\textbf{0.726}} \\
    \textcolor{gray}{East--west wind 7} & \textcolor{gray}{0.551} & \textcolor{gray}{0.475} & \textcolor{gray}{\textbf{0.723}} \\
    \textcolor{gray}{East--west wind 8} & \textcolor{gray}{0.581} & \textcolor{gray}{0.410} & \textcolor{gray}{\textbf{0.676}} \\
    \textcolor{gray}{East--west wind 9} & \textcolor{gray}{\textbf{0.792}} & \textcolor{gray}{0.707} & \textcolor{gray}{0.788} \\
    North--south wind (3D) & 0.772 & 0.524 & \textbf{0.841} \\
    \textcolor{gray}{North--south wind 0} & \textcolor{gray}{0.792} & \textcolor{gray}{0.481} & \textcolor{gray}{\textbf{0.854}} \\
    \textcolor{gray}{North--south wind 1} & \textcolor{gray}{0.673} & \textcolor{gray}{0.429} & \textcolor{gray}{\textbf{0.776}} \\
    \textcolor{gray}{North--south wind 2} & \textcolor{gray}{0.594} & \textcolor{gray}{0.400} & \textcolor{gray}{\textbf{0.730}} \\
    \textcolor{gray}{North--south wind 3} & \textcolor{gray}{0.564} & \textcolor{gray}{0.393} & \textcolor{gray}{\textbf{0.696}} \\
    \textcolor{gray}{North--south wind 4} & \textcolor{gray}{0.573} & \textcolor{gray}{0.405} & \textcolor{gray}{\textbf{0.710}} \\
    \textcolor{gray}{North--south wind 5} & \textcolor{gray}{0.630} & \textcolor{gray}{0.430} & \textcolor{gray}{\textbf{0.764}} \\
    \textcolor{gray}{North--south wind 6} & \textcolor{gray}{0.692} & \textcolor{gray}{0.462} & \textcolor{gray}{\textbf{0.825}} \\
    \textcolor{gray}{North--south wind 7} & \textcolor{gray}{0.784} & \textcolor{gray}{0.502} & \textcolor{gray}{\textbf{0.914}} \\
    \textcolor{gray}{North--south wind 8} & \textcolor{gray}{\textbf{0.956}} & \textcolor{gray}{0.548} & \textcolor{gray}{1.08} \\
    \textcolor{gray}{North--south wind 9} & \textcolor{gray}{1.46} & \textcolor{gray}{1.19} & \textcolor{gray}{\textbf{1.06}} \\
    Absorbed shortwave radiation & 0.982 & \textbf{1.01} & 3.53 \\
    Outgoing longwave radiation & 0.656 & \textbf{0.661} & 2.24 \\
    \bottomrule
  \end{tabular}
  \caption{\emph{Exoclimate emulation:} Spread--skill ratio (SSR). Bold indicates values closer to 1 (better calibrated spread). Rows 0--9 index pressure levels from surface to top for 3D variables; the individual levels are shown in gray. Scores are averaged over five training seeds.}
  \label{tab:exoplanet-AC}
\end{table*}

\end{document}